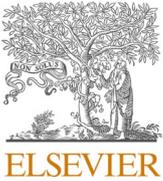

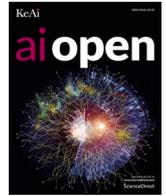

# Discrete and continuous representations and processing in deep learning: Looking forward

Ruben Cartuyvels [*], Graham Spinks [1], Marie-Francine Moens

*KU Leuven, Department of Computer Science, Celestijnenlaan 200A, Leuven, 3001, Belgium*

ABSTRACT

Discrete and continuous representations of content (*e.g.*, of language or images) have interesting properties to be explored for the understanding of or reasoning with this content by machines. This position paper puts forward our opinion on the role of discrete and continuous representations and their processing in the deep learning field. Current neural network models compute continuous-valued data. Information is compressed into dense, distributed embeddings. By stark contrast, humans use discrete symbols in their communication with language. Such symbols represent a compressed version of the world that derives its meaning from shared contextual information. Additionally, human reasoning involves symbol manipulation at a cognitive level, which facilitates abstract reasoning, the composition of knowledge and understanding, generalization and efficient learning. Motivated by these insights, in this paper we argue that combining discrete and continuous representations and their processing will be essential to build systems that exhibit a general form of intelligence. We suggest and discuss several avenues that could improve current neural networks with the inclusion of discrete elements to combine the advantages of both types of representations.

## 1. Introduction

Humans typically communicate with discrete symbols, and natural language is perhaps the most prominent example of such symbolic communication media. Despite the continuous nature of neural signals in the brain and transmission of language via continuous signals of sound, written language is communicated via discrete symbols. One potential underlying reason is the greater information-theoretic signaling reliability and efficiency in communication of symbols versus continuous signals (Shannon, 1948). With deep machine learning taking a central role in artificial intelligence (AI), natural language symbols but also perceptual content are encoded and communicated in the form of continuous representations (sometimes called subsymbolic representations). This is due to the fact that current neural networks are naturally fit to crunch large amounts of real - hence continuous - numbers.

In this paper we will reflect on the role of discrete and continuous representations and processing in the deep learning era and on how discrete and continuous representations could work together, for instance, when humans interact with machines. The central argument in this paper is that both continuous and discrete processing are essential for achieving human-level AI.

The definition of (artificial) intelligence is and has been a debated issue, but Pei Wang made a good attempt: "intelligence is the capacity of an information-processing system to adapt to its environment while operating with insufficient knowledge and resources" (Monett et al., 2020; Wang, 2019). Wang further explains this definition: intelligent systems should hence be able to adapt and learn new skills and acquire new knowledge, they should be able to do so efficiently (insufficient resources) and under uncertainty (insufficient knowledge), they have embodied experience (they experience their environment), are open-ended and should operate in real-time. Other authors add ideas or emphasize different aspects, like social intelligence, which subsumes communication and hence language (Monett et al., 2020). A system can be said to have human-level AI if it has this capacity at least to the same extent that humans have it. Since the subject of this paper is artificial intelligence, and since humans exhibit intelligence, we will sometimes approach artificial intelligence starting from things we know of human intelligence. Even though the fact that humans exhibit intelligence does not imply that human intelligence has to be the only form of intelligence, it is the form of intelligence that we know best and therefore it could provide us with useful insights, teach us about necessary characteristics, and so on, as it has often done before.

Our position, namely that discrete and continuous processing are essential to human-level AI, is motivated by the fact that natural language is spoken by humans around the world, and some even argue that it is innate to humans (Chomsky, 1959, 1965; Chomsky et al., 1988). Natural language is composed of discrete symbols or words. Humans not only take language symbols as input and produce them as output, but they also contemplate language and reason with it. In other words, they internally process symbols (Marcus, 2001). This suggests that the symbols not only have a role in communication, but also that they might affect the thought process of a human in a much more intricate and important way. The *language of thought hypothesis* goes even further and postulates that *all* (human) thinking occurs in a mental language, or 'Mentalese', so that thoughts have syntax, which lends them their compositional structure (Fodor, 1975; Pinker, 1994). Whether this internal symbol processing emerges from purely continuous neural hardware or is supported by discrete or binary components is substance for discussion. Yet, what we do know is that we understand and produce language as part of human intelligence, and that on a functional level we are able to internally process language tokens.

The above position is also motivated by the insight that humans clearly process continuous data. For instance, our feelings are impossible to classify into discrete buckets. We perceive reality as a stream of continuous data consisting of, for example, vision, hearing, taste, olfaction and touch. Our stream of thoughts, or at least the way we perceive it, surely feels continuous as well. Moreover, we have acquired language by grounding it in continuous perceptual signals. This continuity provides large representation power, which is essential when either communicating or reasoning about the real world. The language symbols, as we will argue, are a compression of the meaning they are meant to transmit and by themselves are not sufficient to accommodate understanding. Therefore, the discrete symbols need to be expanded with contextual information, mutual knowledge, expectations, and so on (Shannon and Weaver, 1949; Sperber and Wilson, 1986).

A number of requirements for intelligent systems follow from the above insights. First, intelligent systems need to be able to handle discrete interfaces like language for efficient, effective and reliable communication with humans and potentially with other systems. To represent the real world in an encompassing manner and to reason about it, intelligent systems will inevitably also need continuous representation power. Expanding discrete symbols used in communication with contextual information gives the symbols meaning and makes them understandable. This is shown schematically in Fig. 1: human and machine understanding of communication can adhere to an analogous process. Apart from communication, human reasoning also involves both continuous data and discrete symbol manipulation. As mentioned, reasoning about the real world requires continuous representation power. Internal, cognitive symbol manipulation enables us to reason abstractly, to compose knowledge and understanding, and to generalize and learn efficiently. Although they have great potential, current deep learning systems still fall far short of humans in terms of these qualities. We argue that integrating elements of discrete symbol processing with continuous neural network processing could alleviate some of these shortcomings and hence lead to further progress in AI. This need for reasoning with continuous data, context, spatial and temporal information and potentially symbols and explicit structure is illustrated in Fig. 2.

Starting from the general claim that both continuous and discrete processing are essential for achieving human-level AI, we rely on the following hypotheses:

- Communication with discrete symbols is inherent to human communication, and language is the most obvious example.
- The discrete symbols are a compression of the real world and need to be expanded with contextual information (*e.g.*, visual information) forming continuous representations.
- Reasoning with discrete symbols is sufficient for certain applications, but often, when interacting with the real world, reasoning with continuous representations is more appropriate.
- Humans learn by composing and decomposing representations. Disentangling continuous representations is not well studied yet, although it starts to emerge. In this process discrete representations can play a role to steer the composition process.
- Continuous representations are good in representing meaning, however, we should find ways to link these representations to discrete symbols, the latter being commonly used in the communication with humans.

To support the above hypotheses we have studied the recent literature on representation learning and the debates going on in the field. This leads to a number of insights on how to improve neural network based representations. These insights are evidenced by recent evolutions seen in the literature and demonstrated by our own published research.

The rest of the paper is structured as follows. In section 2, we introduce discrete representations and their properties. Continuous representations in deep learning are explained in section 3. We focus on how they can be learned and on the similarities and differences of the representations across modalities. In section 4, we discuss the limitations of both discrete and continuous processing for reasoning and inference and argue for an integration of both. How we might progress towards such a fruitful integration is the topic of section 5. First, subsection 5.1 considers some problems with the currently widespread brute-force learning of representations from huge datasets. Next, in subsection 5.2, we suggest ideas and discuss current trends in integrating continuous and discrete processing. In section 5.3 we focus on desirable properties of continuous representations that might particularly benefit from discrete elements.

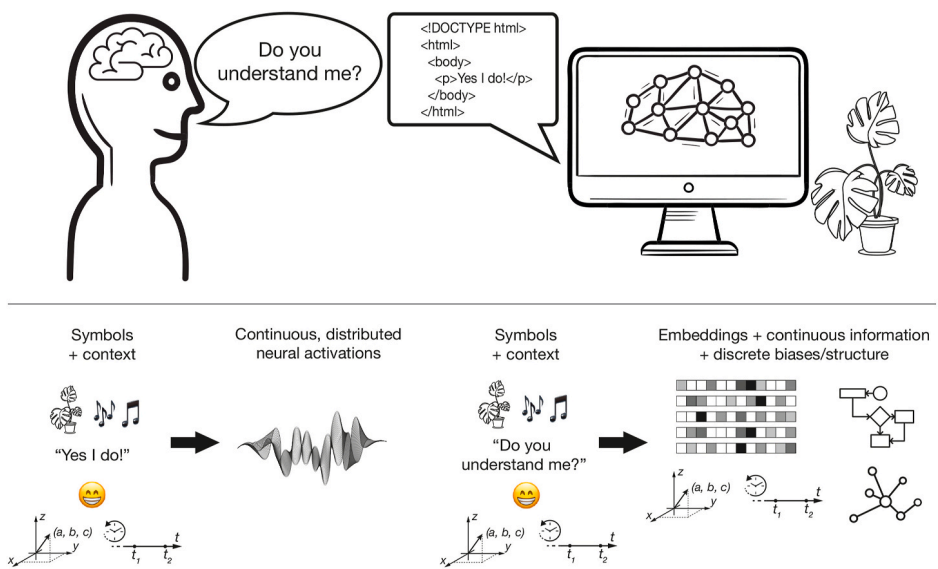

**Fig. 1.** Human language consists of discrete symbols. To communicate with humans and to handle other discrete interfaces, intelligent systems need to be able to handle discrete symbols. These symbols are a compression of the message they are meant to transmit. For the message to be understandable, the symbols need to be expanded with continuous, distributed, contextual information. As we argue in this paper, discrete biases or structure should complement otherwise purely continuous representations. The bottom row of the figure shows how human and machine language interpretation are analogous.

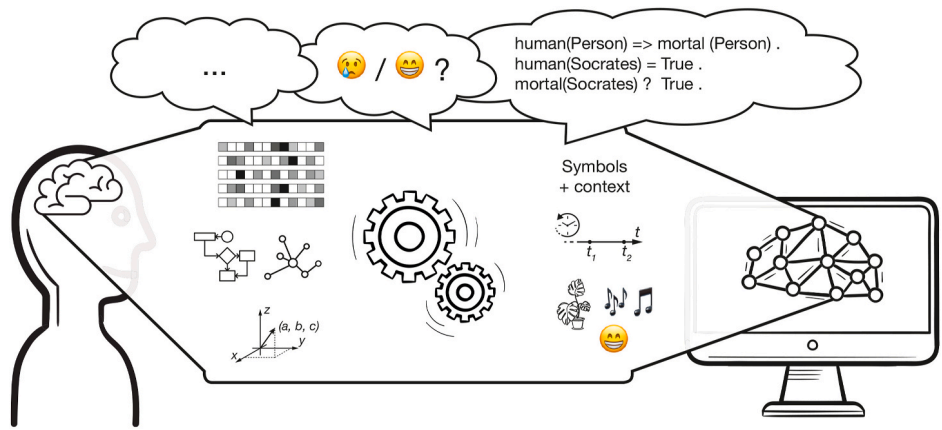

**Fig. 2.** Continuous representations are essential when reasoning about the real world, for both humans and machines. Reasoning also involves other continuous information, like spatial and temporal data. At a functional level (as illustrated by the thought balloons), human reasoning involves symbol manipulation. As we argue, incorporating symbols and discrete structure or biases would bring additional benefits for machine reasoning.

## 2. Discrete representations

### 2.1. Representations

The term 'discrete representation', used throughout this paper, denotes a discretely valued variable that represents some concept, which can take on either a limited or a countably infinite number of distinct values. This definition is purposefully quite general as it remains an open question which type of discrete representation mechanisms might be useful in neural networks. We mostly refer to binary representations, indexes, letters, words or symbols. We use additional 'structure' in neural networks to denote other types of discrete components such as relational structures (graphs, matrices), disentanglement, higher-dimensional arrays, positional enforcing, and so on. When we write about a discrete representation, we generally do not refer to discrete parameters in neural networks such as the amount of layers, the amount of neurons, and so on.

There are **many examples** of humans using discrete symbolic representations in human-human communication and in interaction with machines. Natural language text is composed of strings of word tokens. These word tokens enable fast and exact processing. It is, for instance, straightforward to represent tokens in a computer by their indices in a sorted vocabulary list, that is, with a mapping from the domain of considered tokens to a subset of the natural numbers. Tokens and other discrete representations can also be encoded in *one-hot vectors,* where the presence of a token is indicated by the entry in the vector corresponding to that token having a value of 1, while all other entries have a value of 0. Token equality can then simply be checked by comparing their indices. Moreover, discrete representations have the advantage of being readily interpretable. We can just inspect which word an index refers to. Very often in their communication with machines, humans use controlled language, such as ontologies or symbolic knowledge representations. Controlled language terms can, for instance, be structured in first-order logic, or in triplet structures of subject, object and predicate that populate a knowledge graph.

As another example, mathematics is also described in a language full of symbols, which convey a certain mathematical meaning and interact with an unlimited range of numerical values. Music is encoded in notes and other symbols that are instructive of how the music should sound. Symbolic systems, although often associated with discrete variables (as in formal logic, computer programming), can also involve continuous variables. Evidently, mathematics contains many examples. However, the use of symbols, disregarding the values of the variables they represent, yields discrete parts to the wider system, namely, the symbols themselves, and their syntax or the rules for composing them into valid arrangements.

### 2.2. Properties

Language and other symbolic representations are by nature **compositional**. The syntactical structure of natural language allows for the replacement of words in sentences so that they acquire another meaning. Based on their compositional structure, humans can interpret sentences containing a combination of words that they have never seen before. Controlled languages aim at similar properties, making them powerful tools in automated reasoning. Our experience of the world around us is also compositional in nature. For instance, humans are able to make sense out of visual scenes even if they have never seen a certain configuration of known objects before.

The meaning of symbolic representations is shared by the parties in the communication. For efficiency of the communication, they represent the perceived, or a relevant, world in a compressed format. In other words, the reality of the world is compressed and encoded in a symbolic language. According to the code model of communication from Shannon and Weaver (1949), transmitting a message from a sender to a receiver, including encoding it into a transmittable signal and decoding it back into the message, implies the *mutual knowledge hypothesis.* The hypothesis states that if the receiver is to be sure of recovering the correct interpretation, the one intended by the sender, **every item of contextual information used in interpreting the message must be known mutually by the sender and receiver** (Sperber and Wilson, 1986). A speaker takes into account his and the audience's cognitive environments, and puts in a message just enough symbolic information for the audience **to infer** the intended meaning from the symbols themselves, together with their prior world knowledge, their expectations, the current context and sensory input, and so on.

Symbolic communication is efficient, because the receiver can either decode or infer much richer and more complex meaning than is contained in the symbols themselves. Efficiency can additionally be recognized in Shannon and Weaver's compression, and in Sperber and Wilson's speaker who gives in words (symbols) only the minimal amount of information needed, counting on the audience to fill in the rest. The discrete nature of symbol communication additionally causes the potential for fault tolerance, making use of redundancy[2], which renders the communication more reliable. In addition, recovery in human communication is facilitated by a finite vocabulary of symbols and syntax rules. More importantly, the above model points to the finding that **symbols alone are not sufficient** to capture the communicated content in a comprehensive way. The mutually known or otherwise contextual information is essential for recovery or inference of the correct meaning. This shows the need for richer representations, which is why continuous representations come into play. Fig. 1 shows both the need for symbolic communication and the need to give symbols meaning with context, mutual knowledge, feelings, etc.

### 2.3. Processing

Discrete processing consists of the application of any discrete function to input data. A discrete mathematical function has a domain, and hence a range, consisting only of discrete sets of values. Examples can be found in integer arithmetic, in computer programming languages and in first order logic. AI related examples are functions that convert language tokens into other language tokens or into embeddings. As deep models are currently trained using gradient descent, it is relevant to note that discrete functions are not differentiable (in any subset of their domain). They have no gradients and hence backpropagation is not applicable. An active area of research is concerned with making discrete components amenable to backpropagation by, for instance, replacing discrete functions with continuous approximations (Maddison et al., 2017; Jang et al., 2017; Cuturi et al., 2019; Vlastelica et al., 2020).

## 3. Continuous representations

### 3.1. Representations

In this paper, a continuous representation denotes a continuous, real valued variable, that can take on values in connected regions of $\mathbb{R}^n$. These are generally vectors of real numbers, usually in the Euclidean space[3], that represent a fragment of meaning, and of which the individual dimensions usually have no prespecified meaning[4]. A piece of meaning that such a vector (*embedding* or *distributed representation*) encodes could be anything, like the meaning of a word token, a visual object or scene, or an audio fragment. These embeddings are usually dense, and they are thus meant to represent some meaning in a semantic representation space.

For example, in **Natural Language Processing (NLP)**, tokens are converted to embeddings, to enable semantic processing but also to reduce dimensionality[5]. This conversion to embedding vectors is usually an early step in the processing pipeline. If needed at all, the reconversion to tokens is one of the last steps. Thus, virtually all task-central processing is done in the continuous domain.

Since the domain is continuous and consequently has more degrees of freedom, a single embedding vector can represent more information and carry more meaning by itself than a token. With 'by itself' we mean disregarding the additional information held by communicators to interpret the token. Additionally, its continuous nature allows for nuance to be encoded or for meaning to gradually change. Consider an embedding vector for the token 'dog', of which the first entry encodes fur color. While in words the dog may be either brown or white, the embedding's first entry can take on any real value, corresponding to an unlimited, and gradually changing, range of color hues between brown and white. Importantly, these continuous representations allow for contextual information (*e.g.*, neighboring words, visual or auditory context) to be modelled jointly with the discrete symbols, resulting in a possibly more encompassing representation of reality.

One of the costs of the semantic representation capacity is the loss of interpretability. This is not necessarily the case, but at least currently embedding vectors are typically not readily understandable by humans. Processing becomes inexact and considerably slower, partly because the embedding vectors are typically not sparse. Checking their similarity amounts to computing a distance in a metric space, which is more resource consuming than comparing two integers captured in a one-hot encoding vector. Of course, the exact properties of the continuous representations are dependent of the way they are constructed.

In **computer vision**, activations of final convolutional neural network layers are often used as visual embeddings. These embedding vectors, much like in NLP, represent visual concepts in a semantic space. The represented visual data can be entire images (Vinyals et al., 2015b), regions of images that a model estimates contain an object of interest (Anderson et al., 2018), video clips (Xie et al., 2018), and so on. As will be mentioned in section 3.3, data from numerous other modalities or types can also be converted to continuous representations (such as audio). The same remarks that were made for embedding vectors in NLP also apply to visual and other representations. For instance, their meaning can gradually change, they can encode nuance, and are difficult to interpret by humans.

The use of continuous vectors itself is not free of bias. Using continuous vectors in a Euclidean space or in any other geometric space such as a hyperbolic space (Gulcehre et al., 2019) implicitly assumes that data points, or the relevant aspects of reality, can be meaningfully represented in the respective geometric space, which is not always true.

### 3.2. Processing

Analogous to the discrete case, continuous processing consists of applying continuous functions to input data. A continuous function is one that has an interval as domain and no discontinuities in its value, *i.e.*, abrupt changes in value. Not all functions are either discrete or continuous. They may have discontinuities, while having intervals as domain (and not discrete sets of values). It will be convenient for us, however, to call a function continuous if it is continuous in most of its domain, thus potentially disregarding local discontinuities. We refer the reader to mathematical textbooks on analysis for a treatment of continuity.

---

[2] Error correction codes are examples of redundancy for enhanced reliability in telecommunication. Even though more frequent - but non content bearing - words in natural languages like English are often shorter (Zipf, 1935), yielding higher efficiency, languages contain many redundancies (Shannon and Weaver, 1949), making it, for instance, possible to understand written sentences of which part of the letters are occluded.

[3] Although the vectors may also lie in different spaces, like the hyperbolic space (Gulcehre et al., 2019).

[4] Works exist that attempt to learn disentangled continuous representations with dimensions that correspond to factors of variation in data, but the goal remains far from reached (Locatello et al., 2019).

[5] Although the number of dimensions of an embedding vector may be considerable, usually, and certainly historically, they are still low dimensional compared to one hot encodings of the tokens they represent (Pennington et al. (2014) use 300 dimensional vectors for a vocabulary of 400K to 2M words for GloVe, Devlin et al. (2019) use up to 1K dimensions with a vocabulary size of 30K for BERT, while Brown et al. (2020) use up to 12K dimensions with a vocabulary size of 50K for GPT-3).

In our case, continuous processing will mostly denote (modern) neural network processing. Although neural networks epitomize continuous processing, they inevitably contain discrete elements. The number of token slots in a text transformer (Vaswani et al., 2017), the number of object slots in an object detection transformer (Carion et al., 2020), or the number of nodes in a hidden layer, are discrete parameters. A function can thus be discrete in one variable and continuous in another. For gradient descent to work, the functions implemented by neural networks, as used during training, have to be differentiable functions of at least those parameters that we want to train. That is, the discrete neural network parameters mentioned above cannot be optimized using gradient descent, but the network's weights and biases can. It is worth noting that historically, many neural networks also featured discrete-valued nodes, like (restricted) Boltzmann machines or Hopfield networks (Little, 1974; Hopfield, 1982; Hinton and Sejnowski, 1983; Smolensky, 1986).

Discrete and continuous functions can be composed. The resulting function may be continuous in some variables and parameters, but certainly not in all of them. Take for instance an NLP method that uses language tokens (discrete) as input, converts these to embeddings (continuous), applies a neural network (continuous), and reconverts computed logits into output tokens (discrete). The function implemented by this method, that maps input tokens to output tokens via a number of intermediate steps, is a discrete function, and hence cannot be differentiated. The logits, however, typically are a continuous function of the input embeddings and of the neural network's parameters (not of the input tokens). Gradients of a differentiable loss function that operates on the logits, and not only on the output tokens, can thus be backpropagated to the network's parameters and the embedding vectors.

Modern neural networks are representative of continuous processing. First order logic or symbolic AI models are representative of discrete processing. Many AI models fall between those two poles, for instance, probabilistic graphical models, like the Boltzmann machines mentioned above. Here, we will not elaborate on probabilistic graphical models in general or other AI systems. Instead, we will focus on symbolic and distributed representations in the modern deep learning context[6].

### 3.3. Representations of natural language

In the case of natural language, the unit symbols are words, characters, sentences or documents. The embeddings represent these words (or characters, sentences, documents) in a semantic representation space.

A word embedding can carry meaning beyond the precise meaning of the word it represents. For instance, it can capture the context the word usually appears in (like other words, or perceptual information), it can contain nuance, or it can convey information that would have been implicit, or assumed common knowledge, if transmitted in symbols only. Returning to the communication model of Shannon and Weaver (1949) and the mutual knowledge hypothesis, we could **define a good continuous representation, for general purposes, as one that contains all mutual, contextual information that the sender and receiver both need for encoding and recovering the correct message into and from the signal** (the signal being in this case a symbol or an encoding of it).

### 3.3.1. Learning

The learning of distributed vector representations for words has been studied for a long time (Elman, 1990; Bengio et al., 2003). This learning is often inspired by the *distributional hypothesis*, which states that words that occur in the same contexts tend to have similar meanings (Harris, 1954; Firth, 1957). Following this hypothesis, Mikolov et al. (2013a) and others learned word embeddings by modelling words from their contexts, or contexts from words. They used the weights of a shallow, trained neural network as representations for individual words (Mikolov et al., 2013a,b; Pennington et al., 2014). Here, a context denotes a group of surrounding words, as encountered in a text corpus: The word "fox" might thus be predicted from a context "the fast … catches the mouse", or vice versa. For inference, contexts are disregarded and words are replaced individually by their learned embeddings. These resulting pretrained embeddings proved useful for a wide variety of downstream NLP and multimodal tasks (Frome et al., 2013; Socher et al., 2013; Collell et al., 2017).

While these approaches learn embedding vectors by considering the context in which a word *usually* appears, recent popular representation learning methods also take into account the context in which a word *currently* appears, both during training and inference (Deschacht et al., 2012; Howard and Ruder, 2018; Peters et al., 2018; Devlin et al., 2019; Radford et al., 2019). Starting from individual word embedding vectors, which are context-independent, the representations are contextualized by undergoing changes modulated by the embedding vectors of other words in the current context. Hence, the word "fox" can have two entirely different contextualized representations in two different sentences. This idea, together with the pretraining of big transformer models on massive amounts of text, further improved downstream results by large margins (Howard and Ruder, 2018; Peters et al., 2018; Devlin et al., 2019; Tenney et al., 2019; Raffel et al., 2020; Rosset, 2019; Brown et al., 2020).

The above two embedding schemes are pretrained and provide general-purpose continuous representations that can be finetuned on downstream tasks. Task-specific representations are often learned as well, by starting from randomly initialized vectors and updating them with gradient signals from loss functions of downstream tasks (Sutskever et al., 2014; Vinyals et al., 2015b).

### 3.3.2. Humans

Interestingly, both neurology and cognitive psychology studies suggest that hearing word tokens invokes a similar embedding process in human brains and minds. For instance, neurologists manage to predict activity levels across fine-grained brain regions while subjects listen to stories, from word embeddings of the words in the stories that were trained on text corpora (Mitchell et al., 2008; Huth et al., 2012, 2016; Abnar et al., 2018).

Psychologists have repeatedly observed semantic and associative priming effects (Meyer and Schvaneveldt, 1971; Ferrand and New, 2003), where, simply put, subjects recognize a word like 'cat' more quickly when they have first been shown a related word like 'dog', suggesting that hearing 'dog' already activates brain regions associated with 'cat'. Both words that are semantically similar ('dolphin' and 'whale') and words that have no similar meaning but are often associated ('spider' and 'web') are shown to prime each other. Models for language comprehension that explain these effects have been proposed (Collins and Loftus, 1975; McRae et al., 1997; Landauer and Dumais, 1997) and support concept encoding by distributed representations. While the above studies generally concern single word priming effects, there are also accounts of contextualization of word meaning, where sentences, paragraphs or visual context facilitate comprehension (Bransford and Johnson, 1972; Barclay et al., 1974; Foss, 1982; Stanovich and West, 1983; McClelland et al., 2019). Although current machine language models do not fully mimic what we know of human language understanding (McClelland et al., 2019), we can conclude that there are several elements that the two have in common.

---

[6] However, many neural networks that estimate probabilities actually are probabilistic graphical models, from classifiers to VAEs (Kingma and Welling, 2014). Goodfellow et al. (2016) elaborate on the connection in chapter 16 of their book.

### 3.3.3. Limitations

Distributed language embeddings are considered a compression of the information they contain. Typically, a large vocabulary is mapped to a fixed-length vector where meaning is distributed over various dimensions. Meaning is compressed to the extent that notions about contextual semantics and syntax are distributed among various dimensions (Mikolov et al., 2013a). A downside of such representations, however, is that access to the discrete information is lost. Additionally, it is unclear what the optimal compression dimension should be. Whereas traditional dimensionality-reduction methods, such as singular value decomposition or non-negative matrix factorization, provide a rationale for a choice of dimension size, no such interpretation exists for distributed language embeddings. A variable length representation would in fact be more logical for language. It appears unnecessary that frequently used words with little semantic meaning (e.g., 'the', 'a', 'and', etc.) should have the same length of embedding as more complicated, nuanced words. This limited flexibility in representation length does not extend to discrete representations. As an example, in natural language, frequently used function words are short with an easy pronunciation (Zipf, 1935). Additionally, due to the fixed length of distributed embeddings, neural networks have an enormous amount of parameters, as they need to process all possible combinations of all representation nodes. This is computationally intensive. It would perhaps make more sense to process only relevant meaning dimensions. Attention mechanisms in language models (Bahdanau et al., 2015) address this issue to some extent by focusing on relevant dimensions. Sparsity and structure often found in discrete representations or components could also relieve computational complexity. **We thus argue that discrete representations could compress meaning more sensibly and in a more easily manipulable manner, and hence complement continuous, distributed embeddings.** Some of these aspects are also relevant for visual features, which we discuss in the following subsection.

### 3.4. Representations of visual data

Contrary to natural language, the visual world consists of continuous wavelengths within a visible spectrum. For graphics and imaging applications, this information is usually captured in individual pixels with a quantization of 256 values for red, green and blue color channels. For machine learning applications, these RGB values of the pixels are usually transformed and normalized into a continuous range to obtain a mean of 0 and a standard deviation of 1. This representation then closely corresponds to the continuous nature of vision. As a side note, it is also possible to encode images into a more discrete representation for neural networks. This can be achieved, for example, by clustering RGB pixel values or visual embeddings with k-means. Models traditionally used for language sequences can then be applied to vision problems (Sun et al., 2019; Chen et al., 2020).

### 3.4.1. Learning

It is widely accepted that **locality** is an important characteristic of visual data. To exploit this, convolutional operators are used to perform mappings that take regional information and patterns into account in a bottom-up manner (LeCun et al., 1989; Ciresan et al., 2011). By learning optimal filters, basic characteristics such as lines can thus be combined into more complicated patterns such as squares. When trained for a downstream task, eventually the network learns an embedding containing features from which predictions can be made. This is often done in a classification setting, although many other settings are also possible[7] (Krizhevsky et al., 2012; Anderson et al., 2018; Jean et al.,

2019).

The embedding vectors obtained from the penultimate layer of a fully trained convolutional neural network contain an abstract representation of the original visual input and are frequently used as input features for complementary tasks (Krizhevsky et al., 2012; Razavian et al., 2014; Simonyan and Zisserman, 2015; Xu et al., 2015; He et al., 2016; Anderson et al., 2018; Sung et al., 2018). The idea is that they compactly represent relevant features of the input. Such features comprise patterns that are unique to certain objects, even if the object was never seen during the training of the original network.

### 3.4.2. Lack of understanding/compositionality

This mechanism shows some rough similarities with how humans perceive the world, in the sense that a vast amount of continuous input information is distilled into a compressed understanding. This compression is achieved based on the appearance and characteristics of objects and their relations. However, contrary to humans, such neural networks do not appear to have an adequate understanding yet of what discrete entities are present, nor how they relate to their surroundings. To the contrary, it can be shown that scrambling parts of the image does not impact the classification outcome for deep neural networks (Brendel and Bethge, 2018). Moreover, adding noise to images or removing source tags from images does change neural network predictions, while keeping the images perfectly recognizable for humans[8] (Jo and Bengio, 2017; Lapuschkin et al., 2019). This suggests that neural networks rely mostly on local patterns and textures rather than a fundamental understanding of the content.

The lack of decomposed understanding forms a problem in many cases. A particular type of prize fish would be most accurately classified not by its anatomical features, but by the detection of fingers of the person holding it up for a photograph (Brendel and Bethge, 2018). While this example highlights the ability of continuous embeddings to take the current context and semantics into account, it also shows a failure to assimilate the composition of discrete entities that make up a scene. In contrast, humans are able to distinguish, count, track, identify and represent separately a number of distinct objects featured in their visual input (Xu and Chun, 2009).

This failure inevitably leads to a lack of generalization capabilities outside the setting of the given dataset. Any fish that is held up by a person would likely be classified as that particular prize fish. **Learning better representations that assimilate discrete, compositional information about the world** is thus an important research direction for visual data.

### 3.5. Representations of other modalities

The above techniques can also be used to build continuous representations of other modalities. For instance, given input speech signals, the layer right below the softmax layer of a deep neural architecture trained to estimate text from speech yields a compact representation of the input acoustics (Bengio and Heigold, 2014). Concurrently, word embeddings in an acoustic representation space are learned. Speech representations are pulled by the objective function towards the word embeddings of correct tokens in the joint acoustic representation space, while being pushed away from word embeddings of incorrect tokens. The speech and word embeddings in this work lie in a joint acoustic representation space, rather than in a semantic space, as in most of the works thus far mentioned in this paper.

Continuous representations also play an important role in other audio processing techniques (Graves et al., 2006, 2013; Hannun et al., 2014; Amodei et al., 2016; Chan et al., 2016). To make up for a lack of annotated training data, as in language and vision processing, audio embeddings can also be pretrained in a self-supervised way, to be used afterwards for downstream audio tasks (van den Oord et al., 2018; Jansen et al., 2018; Chung et al., 2019; Schneider et al., 2019). Continuous representations have the possibility to integrate different modalities in an intuitive and natural way, which will be discussed in the next section. An example is the Look, Listen, and Learn system, which produces audio embeddings trained through a self-supervised model of audio-visual correspondence in videos (Cramer et al., 2019).

---

[7] Interestingly, Jean et al. (2019) learn visual embeddings following the distributional hypothesis. Satellite image tiles are presumed to be predictive for other nearby tiles. This illustrates that many different settings are feasible and can still be explored.

[8] Similar results have been published in NLP, e.g., where a model's ability to reason with natural language arguments is shown to be entirely accounted for by exploitation of unintended statistical cues (Niven and Kao, 2019).

We have so far discussed language, vision and audio embeddings, but there are many other modalities that are transformed into representations by deep learning methods. Examples include (but are by no means limited to) action and state embeddings in reinforcement learning (Dulac-Arnold et al., 2015; Chandak et al., 2019), distributed representations for atoms (Schütt et al., 2018), sequences of amino-acids in proteins (Alley et al., 2019), social network posts (Veličković et al., 2019), tactile experience or touch (Gao et al., 2016; Yuan et al., 2017; Calandra et al., 2018) and abstract concepts grounded in olfaction (Kiela et al., 2015).

### 3.6. Joint representations

Recent work has also attempted to combine different modalities into one representation. This trend perhaps mimics the multimodal nature of human reality. Communication often involves multiple modalities (*e.g.*, while a sentence has a particular content, its message is augmented with a particular intonation and body language) and is subject to a concrete context that is experienced through different senses (*e.g.*, the location is seen, the temperature is felt, the music is heard, …).

For many applications, different modalities offer complementary information that can lead to better performance. Combining visual and language features, for example, has been shown to be beneficial in linguistic tasks (Lazaridou et al., 2015; Kiela and Bottou, 2014; Silberer and Lapata, 2014), visual question answering (Su et al., 2019; Lu et al., 2019) and visual commonsense reasoning (Su et al., 2019; Lu et al., 2019; Chen et al., 2019; Li et al., 2020).

#### 3.6.1. Learning

There exist several methods to learn joint representations. Perhaps the most straightforward is to concatenate pretrained features from different modalities (Kiela and Bottou, 2014). These representations have a few downsides. The information from each modality is not aligned, and both visual and language representations need to be available for each example in the dataset. Alternatively, mappings between modalities have been proposed as a method to create informative features (Collell et al., 2017). Another method to obtain joint representations is to learn them jointly, for instance, by extending skip-gram models to several modalities (Lazaridou et al., 2015). More recent works have focused on extending language transformer models to multimodal settings, for instance, text and visual features (Su et al., 2019; Lu et al., 2019; Chen et al., 2019; Li et al., 2020; Tan and Bansal, 2020; Radford et al., 2021), or text and video (Sun et al., 2019). These joint multimodal embedding schemes operate similarly to the equivalent language embedding schemes in that they are pretrained, and they provide general-purpose embeddings for downstream tasks.

#### 3.6.2. Grounding

One motivation to further pursue multimodal representations is the theory that language is grounded in the world and in human actions and experience (Glenberg and Kaschak, 2002; Barsalou, 2008; Bisk et al., 2020). It implies that language and knowledge are acquired through our sensorial and social interaction with the world. Multimodal representations thus partly address this issue by learning concepts and information in a joint manner. The training methodologies, however, mostly still focus on learning distributed representations of entities following the distributional hypothesis. This is done, for example, by extending the unimodal models to include other modalities (Hill and Korhonen, 2014; Lazaridou et al., 2015). While such continuous representations are adept at capturing both visual and textual contexts (Zablocki et al., 2018), a satisfactory grounding in the physical world has not been achieved so far. Multimodal representations have a similar makeup to the unimodal embeddings discussed earlier. They suffer from much of the same weaknesses, and thus they lack the capabilities we seek to represent and manipulate meaning adequately. Language grounding implies that language is understood within a physical context that is compositional in nature and that consists of discrete entities that relate to each other in different ways. **We argue that multimodal representations and embodiment thus also stand to benefit from the advantages of discrete components, structure and biases.**

## 4. Reasoning and inference

Having discussed continuous and discrete representations and processing, and how we might obtain these representations, we will now look at how they play a role in reasoning and inference. Primarily, we see reasoning as the general process of arriving at or moving towards conclusions, driven by arguments or intuition[9] and based on information or knowledge. Inference is defined as the process of making steps in reasoning. These definitions are broad and leave room for the kind of fuzzy reasoning that, for instance, humans employ. Moreover, reasoning often brings to mind formal logic. As such, this reasoning with formal logic can be seen as a more formal subset of the above defined general reasoning.

As humans, we perform all kinds of inference, including but also beyond induction and deduction, and the ability to do so should be a trait of any intelligent system, along with the ability to learn, plan, represent knowledge, perceive, interact, and so on (Russell and Norvig, 2010). That places not only logical but also general reasoning at a central place for the entire field of AI.

### 4.1. Symbol manipulation

The connection between reasoning and symbol manipulation has been made by many philosophers, including Aristotle, Plato and Hobbes (Hobbes, 1914; Łstrokukasiewicz, 1951; Klein, 1989). The language of thought hypothesis mentioned in the introduction also supports the role of symbol manipulation for intelligence, as language consists of symbols, so thinking in language must manipulate symbols (Fodor, 1975; Pinker, 1994).

Early AI approaches were symbolic. They built upon the belief that intelligence could be reduced to pure symbol manipulation[10]. The pillars of these systems are formal logic, symbol manipulating rules and symbolic knowledge bases. Symbolic AI approaches, both early and their modern descendants, are interesting for our discussion, because they are based on discrete representations and processing. Logic, procedures and symbolic knowledge bases all rely heavily on discrete variables, rules and patterns to which variables either adhere or not, and explicit structure. These discretely implemented **symbol manipulation systems excel at effective, efficient abstraction and inference** (Marcus, 2018). Given a set of premises that hold and manipulation rules, a symbolic system deduces consequences that also hold and statements that can never hold. Abstractions, easily represented in symbols, further aid generalization. As mentioned, symbolic knowledge is compositional: the systems have clear prescriptions for how symbols and rules may or may not be composed. That compositionality permits to generalize by combining acquired knowledge as building blocks into new knowledge.

---

[9] Aristotle already drew a distinction between logical reasoning (reason proper) and intuitive reasoning.

[10] See Russell and Norvig (2010); Crevier (1993) for an overview of the history of AI. Although the focus has shifted (Hao, 2020), pure symbol manipulation systems are still an active field of research in AI, and executing symbolic procedures on variables is ubiquitous in computer science in general.

As argued, as humans we also, at least partially, reason with symbols. This symbolic reasoning enables us to operate mentally on concepts of high abstraction levels and to generalize knowledge efficiently. **We should thus require intelligent systems to be able to reason formally and manipulate symbols, at a functional level**. This is also illustrated in Fig. 2. However, humans also incorporate non-discrete data in their reasoning, for instance, estimated probabilities, predictive gut feelings we have about certain future courses, vivid visual memories, and so on. Reality, as we argued, is continuous, and language tokens are a compressed representation of it. The complexity of reality is too high to be accurately and completely encoded in tokens, and semantics get lost in the compression. When reasoning about the real world, tokens are simply too rigid and rules too brittle while the lost semantics, nuance, context, etc., turn out to be essential. Hence, **reasoning about reality will at least need more representation power supplied by continuous representations**. The need for semantic representation power is confirmed by the failure of purely symbolic AI to attain general intelligence. Methods failed to generalize to the real world, outside of either the toy worlds or the abstract tasks for which they were developed. We can thus correct the symbolic AI belief: Intelligence needs symbol manipulation, but cannot simply be reduced to it.

### 4.2. Distributed representations

Deep learning, distributed representations and statistical learning have brought progress towards solving the above problem. Due to their statistical nature, systems automatically learn patterns in data; it is no longer necessary to enter symbolic knowledge or world theories into a system manually. However, while they provide improved semantic representation, intuitive reasoning and perception, **current neural networks leave room for improvement in terms of logical, abstract reasoning, compositionality and interpretability**. We believe that their internal symbol manipulation capability is still limited and hampers their reasoning power.

Promising work has used continuous systems for abstract reasoning tasks (Weston et al., 2016; Santoro et al., 2017; Silver et al., 2017; Hudson and Manning, 2018; Chollet, 2019; Lample and Charton, 2020). One type of reasoning that might especially suit neural learning systems is nonmonotonic reasoning (Antoniou and Williams, 1997), where tentative conclusions can be drawn and later revised in light of further evidence, that is, data (Parisi et al., 2019). But in general, for now, continuous representations still fall far short of symbolic systems in terms of the abstract reasoning they facilitate.

### 4.3. Integration

Because symbol manipulation involving discrete processing and neural networks involving continuous processing both have properties that are essential to intelligence, AI could advance if we consciously integrate ideas from both methodologies. By designing systems that integrate elements from both types of representations and processing, we could take steps towards more human-like or general reasoning and intelligence. The semantic capacity and ability to process perception of continuous systems, as well as their ability to learn from data, could complement the ability of symbolic systems to operate on abstract concepts and to generalize.

Inspiration could be drawn from discrete representations and their processing to relieve some of deep learning's shortcomings. Properties underlying the strengths of symbol systems, such as sparsity and compositionality, could be identified, and one could attempt to transfer these to deep learning systems by imposing biases, adding structure, and so on. The following section 5 will elaborate on this. As a preliminary example, we argue that an intelligent system should be able to manipulate symbols internally at a functional level. Apart from using discrete rule-based components (*e.g.*, as constraints in differentiable objective functions), this need could be met by endowing neural networks with the ability to handle indirection, reference, variables, names and instances. Soft attention schemes take a potential step in that direction, without being symbolic in the classical sense or constrained to strictly discrete representations (Bengio et al., 2019).

In the case of a more direct integration, the counterpart of continuous systems does not have to be symbolic AI: for instance, structured probabilistic methods like Bayesian networks might offer much of the structure, abstraction, sparsity and causality advantages of symbolic AI, while retaining the ability to learn from data (Pearl, 1989; Bengio, 2017; Hudson and Manning, 2019b).

The conclusion from this section is that reasoning only with discrete representations will not be sufficient for systems to acquire a general type of intelligence. Rather, the use of continuous representations could offer better semantic representation for some aspects of reality. Such **continuous representations could be integrated with symbolic, structured or other discrete processing methodologies**, which offer a greater potential for efficient and effective abstraction and generalization.

## 5. Improving neural network representations

We have argued so far that intelligent systems need rich semantics like the ones offered by distributed representations, but that current continuous representations and methods still have shortcomings. We have further put forward two reasons why intelligent systems need discrete components: first, to communicate with humans, and second, because they might help tackle some of the shortcomings of continuous systems (*e.g.*, compositionality or the functional ability to manipulate symbols).

In this section we will have a further look at what is lacking in current neural network representations and how they can be enhanced. We will discuss some concrete methods to combine discrete and continuous representations and their processing. Additionally, we will have a look at compositionality and generalization, and how they could be improved by infusing more structure.

We do not intend to provide an exhaustive list of issues, examples, suggestions and possible ways to improve representations or processing. As discrete and continuous representations and processing, and the interaction between them, are omnipresent in contemporary AI, enumerating all relevant research would be infeasible. Instead, we focus on what we deem are important insights and offer some examples from the subfields where our own expertise lies.

Finally, we note that disagreements between symbolic AI and connectionism and statistical learning in general have occurred recurrently throughout the history of AI, as shown by Minsky and Papert (1969); Marcus (2018). Alongside those tensions, there have also been calls for the combination of elements of both fields (Fodor et al., 1988; Minsky, 1991; Besold et al., 2017; Lake et al., 2017; Marcus, 2020), and correspondingly, there have already been many attempts to do so (Sun and Bookman, 1994; d'Avila Garcez et al., 2009; Manhaeve et al., 2018; Yi et al., 2018). However, the message that we want to convey with this discussion, as summarized by the last sentences of section 4 and the first sentences of this section, is more general than a call for the integration of classical AI with connectionism. Additionally, we hope to shed light on this matter from the viewpoints of the most recent trends in deep learning.

### 5.1. The interpolating brain

With respect to the nature of human intelligence, there exists a train of thought that assumes that the brain applies brute-force fitting of data (Hasson et al., 2020). In this view the brain is essentially an over-parameterized function that learns general and useful interpolations of observations from the world, very similar to neural networks. **They argue that, when the sampling of the input space for a dataset is dense enough, interpolation between these input data can provide for adequate generalization, and generalizing by extrapolation is not required.** Models of increasing capacities should be able to parse and interpolate better between data. In this view, the focus should not lie on clever engineering or extensive pre-processing, but rather on the use of general models with minimal inductive bias. With enough computing power and data, a better solution can be interpolated than can be hand designed. Many of the recent language models, trained on massive text datasets, are congruent with this line of thinking (Vaswani et al., 2017; Radford et al., 2018, 2019; Devlin et al., 2019; Brown et al., 2020; Lu et al., 2021). Text is given as input to models as raw as possible, preserving all conjugations and different word forms or synonyms, instead of being, *e.g.*, lemmatized or grammatically annotated. Transformers function well as workhorses of semantic processing, because they make maximal use of meaning in sentences by letting every word influence all other words in ways dependent on the input itself, thus in a way maximizing intra-sentence contextualization. Without explicit structure, the language models used for processing natural language are sufficiently powerful to induce the structure or grammar rules implicitly contained in language. Current methods appear to be powerful enough to at least handle sentence-level grammar consistently (Conneau et al., 2018; Goldberg, 2019; Hewitt and Manning, 2019), and their representations have already proven useful in a wide range of tasks (Brown et al., 2020; Lu et al., 2021).

#### 5.1.1. Pitfalls

Despite the deep learning community's great confidence in current continuous processing by transformers or other neural networks, there are a few pitfalls. **Theoretically it might be possible to encode any discrete information into continuous embeddings. But current embedding approaches do not yet flawlessly capture all relational, temporal, spatial or other structure that is found in the real world** (Hudson and Manning, 2019a; Liu et al., 2020; Agarwal and Mangal, 2020; Conneau et al., 2018). We believe that this is one of the shortcomings that cause some challenging AI tasks, beyond, for instance, basic natural language understanding (NLU) or image classification benchmarks, to remain far from solved. As an example of a setting where more structure might help, current models struggle to remain on-topic when generating longer discourses (Brown et al., 2020). In visual question answering, previous progress can be partly attributed to exploiting biases in datasets (Johnson et al., 2017a; Hudson and Manning, 2019a). On a more balanced and challenging dataset, current models perform poorly, even though the best models make use of structure for compositional reasoning (Hudson and Manning, 2019b). The Abstraction and Reasoning Corpus requires models to make abstractions and generalize from very few examples, which both current discrete and continuous models fail at[11] (Chollet, 2019). Another consequent shortcoming is that current neural models have no effective way to distinguish causation from correlation (Marcus, 2018; Lake et al., 2017). Feature co-occurrences in training examples are exploited, but underlying causal relations are not modelled. This has the adverse effect that models exploit *unintended* clues in data or "shortcuts" (Lapuschkin et al., 2019; Niven and Kao, 2019; Geirhos et al., 2020). In computer vision, it is not straightforward to represent distinct entities when analyzing images. Features in neural networks typically indicate the presence of a particular characteristic but do not indicate how many or which instances exhibit those features. It is thus not straightforward to identify object instances from the same category nor how different instances interact (Liu et al., 2020; Agarwal and Mangal, 2020). As mentioned in section 3.2, current vision networks also have limited understanding of the 3D world as well as what objects are and how they relate to each other (Agarwal and Mangal, 2020; Tang et al., 2020).

#### 5.1.2. Beyond interpolation: the composing brain

In response to the above pitfalls, a different train of thought has also formed in the deep learning community. **The idea is that the brain employs a mechanism beyond interpolation or brute-force learning of correlations in data, which allows, *inter alia*, for causal modelling and compositionality.**

The ability to compose elements and their relations together into more complex elements, and vice versa, to decompose complex elements into simpler elements, is regarded as key to human intelligence (Battaglia et al., 2018). The ability to compose induces so-called *combinatorial generalization*, that is, small sets of known building blocks can be dynamically combined into limitless compositions, allowing to make "*infinite use of finite means*" (von Humboldt et al., 1999). By combining simpler, known elements into a combinatorial number of unknown compositions, an intelligent system can generalize to unknown situations. Given their combinatorial growth, it would never be possible to encounter all of these situations during training, which means generalization is the only way of dealing with them. Combinatorial generalization is what systems need to perform open-ended inference. Combinatorial generalization based on discrete symbols is straightforward, but when dealing with continuous representations, it is still an open problem (Marcus, 2018).

Compositionality relates closely to sample efficiency. On tasks where computers have outperformed humans, the necessary number of training points for doing so is generally huge, often more than humans would ever be able to process in one lifespan (Silver et al., 2017). If an intelligent system that has learned smaller building blocks could compose these blocks indefinitely and draw inferences from the compositions, it would not need to be explicitly taught about the combinatorial amounts of compositions. Intelligent systems not only have to generalize and perform well, but they should also be able to do so efficiently, both in time and in the amount of data needed. Chollet (2019) even bases his measure of intelligence on efficiency of skill acquisition. Therefore compositionality has the potential to improve both generalization and sample efficiency drastically.

**With respect to language**, for example, **it is simply intractable to learn the meaning of all possible sentences or all possible discourses separately**. The closest we might get is to train on immense datasets, as demonstrated by some of the recent language models we discussed in the beginning of this section, such as GPT-3 (Brown et al., 2020). As we already mentioned, these models do exhibit some compositional properties: They can generally generate syntactically correct sentences, which requires an 'understanding' of the rules of composing language: syntax. It is unclear, however, to which extent the network relies on pattern memorization rather than compositionality. Several studies of compositionality of language models have recently been published (Lake and Baroni, 2018; Loula et al., 2018; Hupkes et al., 2020). The main outcome is that current models do not yet generalize by composition to a satisfying extent, although the tested models do not perform poorly on every aspect of compositionality. Instead, humans certainly do understand language by learning the meaning of words and composing more elaborate meanings with sentences and syntactical structures. That suggests that, in order to progress towards human-level language understanding, we need to improve the composing abilities of language models.

We argue that this mechanism beyond interpolation could perhaps be replicated, or approximated, in neural networks by adding discrete representation power, processing and/or additional structure. The following two subsections will give some suggestions of how that could work.

---

[11] The best scoring submission in a 3-months Kaggle competition got an error rate of 79 % (Kaggle).

*5.2. Integrating discrete and continuous representations and processing*

*5.2.1. Discrete intermediates*

One relevant question is whether a discrete representation could serve as a useful sole, intermediate representation in a processing pipeline. In our view, if meaning is important and the processing concerns semantics, this is unlikely to be beneficial, since nuance, implicit information and context would be lost (cfr. the communication models discussed in section 2). **The discrete semantic intermediate would become incapable of retaining sufficient meaning.** As was already discussed, continuous representations in the form of embedding vectors are more fit than tokens as meaning-carriers. Below, we discuss some more promising ways of integrating discrete and continuous representations.

*5.2.2. A step towards integration: attention mechanisms*

A technique that could be seen as a step towards integrating discrete and continuous representations, and that is widely used today, is the attention mechanism. An attention mechanism typically outputs a weighted sum over a discrete number of input features (mostly distributed representations), where the weights are determined by a softmax over inner products of corresponding keys and a query, which are usually also distributed representations[12]. Hence, attention allows for a soft selection or combination of input features, modulated by a query. For a particular outcome, the attention weights can give an indication of the importance of an input feature (word embedding, pixel, visual feature, …), providing a form of interpretability of the network, that is, attribution. Attention is used in systems handling different modalities, like language (Bahdanau et al., 2015; Vaswani et al., 2017) or images (Child et al., 2019; Chen et al., 2020). The query can also be taken from another modality, to fuse multimodal information. Accordingly, attention is often used in cross-modal applications such as text-to-image generation (Xu et al., 2018), visual question answering (Yu et al., 2019), or image captioning (Xu et al., 2015; Anderson et al., 2018).

Apart from yielding **a soft, continuous version of an otherwise hard, discrete select operation**, attention has some further interesting properties. Attention allows to **dynamically build outputs by choosing or combining features from a set of inputs, which provides some combinatorial power to build new meaning from smaller parts**, a desirable feature for generalization (see also section 5.3). By operating on sets of a discrete number of distinct representations, or "slots", attention turns neural networks from vector processors into set processors. The ability to represent and compose distinct parts potentially leads to disentangled representations of compositions. Interestingly, the soft select can be used not just to combine given input features, but also to approximate other hard decisions in a soft manner, like what submodule to execute or which memory location to write to (Graves et al., 2016; Hu et al., 2018). A final insight is, as we already briefly suggested, that attention provides a level of **indirection between a reference to an output variable slot and a number of input variables** (Vinyals et al., 2015a; Hudson and Manning, 2018; Carion et al., 2020; Locatello et al., 2020; Bengio et al., 2019). Furthermore, the input variables can be seen as named (by their keys), and so can the output variable (by the query). These are properties reminiscent of symbolic systems such as programming languages, but the attention operation remains fully differentiable.

*5.2.3. Explainability*

A generally intelligent system should be able to explain its thought process and its decisions. Also for a better understanding of neural models, as well as for debugging and engineering, more transparent processing would be of value. We just mentioned that attention weights offer a limited form of explanations, by attributing relative importance in the realization of a given outcome to the different inputs. However, **the explanation of reasoning and decisions of course constitutes a form of communication between machines and humans, and hence, inevitably, truly explainable systems will require symbols** and therefore discrete representations. Explainability in neural networks constitutes a broad research subfield by itself, and for a thorough discussion the reader can refer to Marcinkevics and Vogt (2020).

*5.2.4. Other ways to combine continuous and discrete processing*

Apart from attention, there are a number of other techniques that are commonly used to make otherwise discrete, non-differentiable processing amenable to gradient based learning. For instance, policy gradients (Williams, 1992; Sutton et al., 1999) are frequently used to backpropagate through discrete decisions (Xu et al., 2015; Hu et al., 2017). Reparameterization tricks, such as the Gumbel-Softmax distribution, offer a differentiable mode of selecting one of several distinct categories in a sampling setting (Jang et al., 2017; Maddison et al., 2017; Kipf et al., 2018). Without doubt the most prevalent method to convert continuous representations into discrete output is by modelling a categorical probability distribution over the discrete elements, parameterized by the continuous input (Bengio et al., 2003). The reverse mode is converting a discrete element like a word token in language or an action in reinforcement learning to a distributed representation. This is typically solved by extracting the right row or column, which is the wanted embedding, from an embedding matrix using the token index as index.

*5.2.5. Indexing systems and memory*

Indexing systems in information retrieval traditionally make use of discrete symbols for efficient and exact matching, but can benefit from continuous components and semantic matching, as suggested by Wei and Croft (2006); Mitra et al. (2017); Zamani et al. (2018). By imposing sparsity to continuous embeddings, Zamani et al. (2018) find a middle ground in terms of continuous and discrete qualities. The resulting representations exhibit improved interpretability and generalization, while at the same time they provide contextualized meaning. Interestingly, there is evidence that the brain utilizes a similar mechanism in memory retrieval (Polyn et al., 2009; Yonelinas et al., 2019; McClelland et al., 2020). Concepts or objects are encoded not only through their semantic properties but also through the context they appear in. This means that concepts can be retrieved from memory in different manners. The first manner is based on the particular characteristics that an object has, the second manner retrieves the concept based on contextual information. Humans are often reminded of a concept by seeing associated content, rather than the concept itself. As a concrete example, the human mind draws from past experiences when encountering an ellipsis in a text, or when content is obfuscated in images.

Likewise, continuous systems can benefit from (partially) discrete indexing components like an explicitly or sparsely indexed memory. A series of work has explored extending neural networks with external memories or tapes, which the network can read from and write to (Graves et al., 2014; Weston et al., 2015; Sukhbaatar et al., 2015; Graves et al., 2016). Typically, in order to keep the system end-to-end differentiable, reads and writes are modelled by soft attention mechanisms (see above), so every memory location has a non-zero contribution to read results, and every location is accessed during the write operation. This leads to escalating time and memory footprints. Therefore, Rae et al. (2016) make these memory networks more scalable by making the attention weights **sparse**, in a similar vein to Zamani et al. (2018). Apart from the efficiency argument, Zaremba and Sutskever (2015) note that **model capabilities could be further extended by learning how to interact with discrete interfaces** (such as discrete memories, input and output), **since many existing real-world interfaces are in fact discrete** (databases, programs, search engines and, interestingly, also natural language). As we mentioned already, intelligent systems should be able to communicate with humans. Since human language is discrete, machines need to be able to interact with discrete interfaces like language. Gülçehre et al. (2018) test neural networks extended with both continuous and discrete access memories, and find that discrete memories often outperform their continuous analogues. Xu et al. (2015) reach a similar conclusion when comparing soft and hard attention over image regions for caption generation.

---

[12] We refer the reader to Bahdanau et al. (2015) or Vaswani et al. (2017) for a more complete description of attention.

### 5.2.6. Top-down and bottom-up

Inducing theories from observations and deducing predictions about observations from abstract theories, are both goals for intelligent systems. To achieve both, systems could combine two different learning approaches. The first, which is bottom-up, is the widely used mechanism to minimize objectives from very large training sets, which has proven its value in many semantics tasks that operate on continuous representations. The second, which is top-down, would allow to reason about concrete examples starting from abstract information, an ability that is typically more easily achieved with discrete or structured approaches.

One possible approach to achieve top-down learning from limited data is to use discrete representations of concepts and to apply cardinal minimization to an algebraic semilattice structure. This can be done in a parameter-free setting with simple heuristics without overfitting on the data (Martin-Maroto and de Polavieja, 2018). Various other top-down and bottom-up integrations could be conceived. Manhaeve et al. (2018), for instance, extend a probabilistic variant of a logic programming language to integrate with and take output from low-level neural networks, where the neural network's parameters and those of the probabilistic program can be learned. However, the structure from the high-level program remains fixed, hence it is not learnable, and it only operates on probabilities and symbolic variables, not on rich distributed representations. As another example, Hudson and Manning (2019b), drawing inspiration from Bengio (2017), predict a probabilistic dependency graph for VQA that represents scenes in which nodes and attributes have distributed representations.

Generative models could alternatively instantiate the top-down part, implementing a mapping from abstract (e.g., class labels, text or latent representations) to concrete samples. Nair et al. (2018) provide an example: they use a generative network to generate goals, represented by disentangled embeddings, which they use as objectives for self-supervised learning.

Such ideas, including the notable role for top-down learning, have received less attention in the deep learning literature, but have enormous potential when combined with neural network representations. Yet they are at the essence of what we want to achieve by combining continuous and discrete representations/processing. **With integrated top-down and bottom-up representation mechanisms, bottom-up learned concepts could be combined and composed within particular top-down contexts. As such, ideas could be generalized outside the contexts in which they were learned.** This brings us to the related topic of compositionality and generalization. As we argued, they are two aspects in which current deep learning still falls short.

### 5.3. Compositionality and generalization

As we explained in section 5.1, composing and decomposing concepts is an essential trait of intelligence, as it enables combinatorial generalization, and as a corollary, could drastically reduce the problematic sample complexities of current systems. Compositionality relates to our discrete and continuous discussion, since language and other symbol systems comprising discrete representations are often inherently compositional, as discussed in section 2. In this section, we put forth some promising ideas and lines of research that aim at enhancing compositionality in deep learning systems.

### 5.3.1. Structure

Structure is abundant in the world, and can come in many shapes: temporal, spatial, compositional, relational, and so on. It is essential for humans to understand both the world and language, to assign meaning to its constituents correctly and to further process it. Recognizing and understanding structure in otherwise chaotic observations seems a valuable aid in simulating any sort of intelligence.

Continuous representations of content that correctly capture a specific kind of structure can already be obtained by training models to extract that structure. Temporal structure can be learned by ordering events in text onto an absolute timeline (Leeuwenberg and Moens, 2018; Leeuwenberg and Moens, 2020). Spatial structure can be extracted from textual relationships (Collell et al., 2018). In addition, Hewitt and Manning (2019) and Goldberg (2019) show that embeddings computed by BERT encode syntactic structure (Devlin et al., 2019). In contrast to the prior works about temporal and spatial structure, BERT was not trained explicitly to extract syntactic structure. Yet, **how to capture the same variety of structure as humans from diverse observations in purely continuous embeddings remains unsolved.** Part of the difficulty originates from the fact that current methods obtain the representations by optimizing an objective function related to a targeted downstream task. But such a loss is determined by many more factors than the ability of the obtained representations to capture structure, and thus might provide a noisy signal for the embedding method if the goal is to explicitly represent structure.

These works show that structure, which is often characterized by discrete elements, can already be implicitly captured to some extent in continuous representations. This implicit structure could benefit compositionality, hence we believe further exploring it could prove worthwhile.

### 5.3.2. Disentanglement

While the concrete techniques to improve compositionality of continuous representations are not yet clearly established, one avenue is to improve the disentanglement and interpretation of the neural models. Recurrent neural networks, for example, can be shown to achieve basic compositionality on toy examples (Kharitonov and Chaabouni, 2020). They add an inductive bias towards sequential structure, which can be suited for temporally or otherwise sequentially structured data. It is therefore not unthinkable that more structure could be added to continuous representations by construction.

A possible solution is to create disentanglement in continuous representations of existing methods. As an example, representations in variational autoencoders can be encouraged towards disentanglement by imposing constraints (Higgins et al., 2017; Kim and Mnih, 2018). Similarly, a reconstruction error based on a Principal Component Analysis can drive structured features toward disentanglement in an autoencoder (Pandey et al., 2020). Additionally, structure can be imposed in a learning process by comparing discrete class labels with randomly chosen contextual dimensions (Spinks and Moens, 2020). Such representations offer interesting qualities, as they can be decomposed into meaningful components, modified and recomposed into a continuous representation. Disentanglement is thus in general a desirable property for compositionality.

The term 'disentanglement' is often used to denote dimensions within one embedding that represent disentangled factors of variation. In a broader sense, **observations could also be represented in a disentangled manner by encoding different constituent factors or parts with their own continuous representation.** Changing or substituting one constituent of the observation, while keeping the rest equal, yields a localized change in the corresponding representation; only the representation of that constituent changes, which is the goal of disentanglement. This idea is gaining importance, inter alia, in computer vision. Before, single representations of an entire image were used as input to downstream tasks (Vinyals et al., 2015b; Xu et al., 2015). Now, it is becoming standard to condition on an essentially disentangled representation of an image consisting of a set of representations of its most salient constituent objects (Anderson et al., 2018; Burgess et al., 2019; Watters et al., 2019; Engelcke et al., 2020). This form of disentanglement fits well with the attention operation, that selects or merges information from the disentangled representation. When we additionally model relations between these constituents (that together form a set with a discrete number of elements), the structure we end up with is a graph.

Disentanglement could also play a role in modelling causality. If causal factors are to have a delineated and recognizable effect, corresponding changes in distributions, for instance, of input, outcomes or output variables, should be localized (Bengio et al., 2019). This would be hard to achieve if underlying factors, like agents or objects in a scene, are completely entangled in a representation.

### 5.3.3. Graphs

Other work suggests that for more complex tasks more structure in the neural architectures is required than disentangled (sets of) representations (McCoy et al., 2020). That structure could be derived from the inherent compositionality of graphs, as advocated by Battaglia et al. (2018). Graphs and graph neural networks offer a structural bias towards relational information, similarly to convolutional neural networks that impose a locality bias (Scarselli et al., 2008; Zhou et al., 2018). Given the relational structure of many real-world systems, this structural bias might be a desirable property. **The use of continuous representations as attributes in discretely structured graph representations gives a form of hybrid representation and adds structure to neural network processing**. For instance, recent research shows that symbolic models of the world can successfully be deduced from observations (Cranmer et al., 2020). By encouraging sparseness in graph neural networks, the authors manage to extract symbolic information about physical processes. This research direction aligns with our views, as it combines the advantages of both continuous representations and symbolic components.

Interestingly, applying self-attention as in transformers (Vaswani et al., 2017) to a set of features can conceptually be seen as message passing between nodes in a fully connected graph, an analogy proposed by Veličković et al. (2018). In this view, (self-)attention can be used to construct arbitrary directed graphs from input nodes in a soft manner. Graphs, much like sets as used in attention, keep factors disentangled compared to representing everything at once (in a single vector), because nodes and edges are separately represented. This allows to model relational and causal systems, since localized changes can be computed resulting from cause and effect events (Kipf et al., 2020).

Graphs have also found use in other applications, for instance, to generate images from language, as shown by Johnson et al. (2018). The method they propose converts a scene graph, which is a graph representation of a sentence, into an image. They show that with the explicit structure of the graph, they are able to generate more semantically correct images than a method that converts raw text into images (Zhang et al., 2017).

Finally, another representation related to graphs that appears very promising is a *factor graph*, that represents a factorization of a function. Any factor in such a graph creates a direct dependency between a set of variables. Such structures can potentially be used in representation learning tasks for high-level concepts, which would be essential for higher-level, abstract reasoning. A sparse factor graph would then model the joint distribution over high-level concepts. It has been suggested that such an approach mimics the functioning of human consciousness (Bengio, 2017). This idea shows promising results on a challenging visual question-answering dataset, although still performing much below human level (Hudson and Manning, 2019b).

### 5.3.4. Approximate inference and guiding search

Earlier in this paper, we and other authors have argued that more structured, sparser and probabilistic models, like hierarchical Bayesian models or hidden Markov models, might offer some of the qualities we seek to add to deep learning. They would still be trained from data and would remain compatible with distributed representations but could also enable higher-level, conscious processing, like abstract and causal reasoning (Lake et al., 2015; Bengio, 2017; Lake et al., 2017; Hudson and Manning, 2019b). However, these types of models often suffer from intractable exact inference. For instance, exact inference in Bayesian nets (Pearl, 1989) is proven to be NP-hard (Cooper, 1990). Approximate, Monte Carlo-based inference methods are used to cope with this factor, but in vast hypothesis spaces, with only few good solutions, inference remains difficult. As Lake et al. (2017) suggest[13], humans are able to come up quickly with a small set of plausible solutions when faced with difficult problems or questions, hinting towards an intuitive or subconscious mechanism to select hypotheses to test. This makes humans efficient at inference and search. Along these lines, **one promising idea is to use the intuitive processing of neural networks for approximate, amortized inference in structured probabilistic models or for inference in settings characterized by vast hypothesis spaces with few solutions.**

This idea has already been explored by, for instance Heess et al. (2013); Eslami et al. (2016); Yoon et al. (2018); Satorras et al. (2019); Satorras and Welling (2020), but many possibilities remain for further work. Kingma and Welling (2014); Rezende et al. (2014) use neural networks for amortized inference, which learns an efficient mapping from samples to proposal distributions in probabilistic models. Y. Bengio makes a very similar argument in a recent talk (Bengio et al., 2019): Neural networks could guide efficient search in reasoning. The overwhelming success of DeepMind's Go-playing systems (Silver et al., 2016, 2017) is also based on this principle. Instead of expanding random nodes in their Monte Carlo tree search, which would inevitably fail the system given the immense search space of Go, they select nodes predicted by a neural network. Since the search tree is a discrete structure with discrete states, by doing so, they make an interesting case for integrating discrete and continuous processing.

### 5.3.5. Generative compositional models

Progress has recently been made in **guiding generative processes, implemented by neural networks, with discrete input**. Xu et al. (2018); Pavllo et al. (2020b), for instance, generate realistically looking images, conditioned on natural language descriptions. Vedantam et al. (2018) achieve a similar effect but, they condition on sets of labels, and they incrementally refine the image by adding labels, going from abstract to concrete labels. Although there is still room for improvement, changing one word in the conditioning description only changes the corresponding parts in the output image; hence, the generation can be compositionally controlled. Crucially, **the input text has compositional meaning, and the output image adheres to this composed meaning, entailing that the intermediate, continuous representations in the neural networks manage to capture compositionality** to an extent. Pavllo et al. (2020b) generate objects in 3D by predicting both mesh and texture, with neural networks and continuous, distributed representations. Forcing the generation process to be congruent with the 3D space naturally disentangles the shape of generated objects from their pose, facilitating a better semantic alignment between the input text and visual output.

This direction of research has great potential. While currently this mainly works for rather simple images, it could be extended towards complex scenes and videos (Pavllo et al., 2020a).

### 5.3.6. Composing modularized computations

Another particularly relevant series of methods aims to learn composable neural module networks for visual reasoning and visual question answering. Each of these modules specializes in one subtask, and they can be composed into complex, deep networks (Andreas et al., 2016b). The composition depends on what computations are required for answering a given natural language question, and in what order. This has the advantage that not every type of question needs to be learned from the beginning. Rather, the building blocks have to be trained, and one should learn how to compose these blocks. If those two skills can be acquired, the model can generalize to new types of questions by composing the learned blocks into new compositions. A series of studies start from this idea and improve upon the method in various ways (Andreas et al., 2016a; Hu et al., 2017; Johnson et al., 2017b; Hu et al., 2018; Mascharka et al., 2018; Yi et al., 2018; Mao et al., 2019). Most of these methods are to some extent interpretable. Knowing which modules are executed and what input they attend to is informative for an outcome. The composable, specialized module networks can be contrasted to Hudson and Manning (2018)'s model for visual question answering, which repeatedly applies one complex general-purpose block to focus sequentially on different image parts and perform different operations. Although the above works apply the idea of composing neural modules to reasoning about visual data based on language queries, we believe that it could prove useful in a broad range of other settings.

---

[13] We refer to the paper of Lake et al. (2017) for a more thorough explanation and motivation.

From the above we conclude that it is beneficial to explore a variety of approaches that would better mimic how the human brain processes both rich continuous information as well as discrete language. As discussed above, **desirable properties, such as generalization and compositionality, can be achieved with hybrid discrete-continuous representations or structurally informed models**.

## 6. Conclusions

Current AI systems still fall short of reaching real intelligence. In this paper we have postulated that an important cause is the way that machines represent natural language and other modalities and reason with these. Traditional symbolic representations, such as the discrete symbols found in language, are important in human-human and human-computer communication. However, they only provide a compressed representation of the real world and fail in capturing contextual information, which is needed in human-like understanding and reasoning. Continuous representations succeed in capturing contextual information, but they lack a transparent structure that would facilitate composition and generalization. As a result, they are learned by a brute force exploitation of huge data sets. Hence, there is a need to study the value of discrete and continuous representations with a view to their potential integration.

This study is realized at a time of an ongoing debate among prominent AI researchers (Bengio et al., 2019). Some deep learning adepts argue that symbolic processing should emerge from continuous processing, which means that intelligent systems would eventually not need explicit discrete components. Others dispute this position and argue that purely continuous systems will never be able to reach intelligence by themselves and will always need symbolic higher-level reasoning components. Careful analysis of the different arguments found in the literature and our own published research has led us to conclude that we need novel representations that integrate discrete and continuous components in order to simulate human-like understanding of language and other media. We drew inspiration from human intelligence, because human intelligence has both a continuous, reality processing, perceptual aspect and a symbolic aspect that reasons and processes language tokens.

These findings open many venues for future research on topics such as how to fuse discrete and continuous processing in neural networks, how to integrate structure into the representations, how to impose sparsity to continuous representations, how to integrate general human-like reasoning, how to deal with top-down and bottom-up integration of representations and processing, how to enforce compositionality, and how to explain the resulting representations. We firmly believe that solutions to the above questions will advance the state of the art in AI, lead to many desirable properties including efficiency, reliability and explainability and enable improved reasoning and generalization capabilities.

## Funding

The research leading to this paper received funding from the Research Foundation – Flanders (FWO) under Grant Agreement No. G078618N and from the European Research Council (ERC) under Grant Agreement No. 788506.

## References

Abnar, S., Ahmed, R., Mijnheer, M., Zuidema, W., 2018. Experiential, distributional and dependency-based word embeddings have complementary roles in decoding brain activity. In: Proceedings of the 8th Workshop on Cognitive Modeling and Computational Linguistics (CMCL 2018).

Agarwal, A., Mangal, A., 2020. Visual Relationship Detection Using Scene Graphs: A Survey. CoRR abs/2005.08045.

Alley, E., Khimulya, G., Biswas, S., AlQuraishi, M., Church, G., 2019. Unified rational protein engineering with sequence-based deep representation learning. Nat. Methods 16, 1315–1322.

Amodei, D., Ananthanarayanan, S., Anubhai, R., Bai, J., Battenberg, E., Case, C., Casper, J., Catanzaro, B., Chen, J., Chrzanowski, M., Coates, A., Diamos, G., Elsen, E., Engel, J.H., Fan, L., Fougner, C., Hannun, A.Y., Jun, B., Han, T., LeGresley, P., Li, X., Lin, L., Narang, S., Ng, A.Y., Ozair, S., Prenger, R., Qian, S., Raiman, J., Satheesh, S., Seetapun, D., Sengupta, S., Wang, C., Wang, Y., Wang, Z., Xiao, B., Xie, Y., Yogatama, D., Zhan, J., Zhu, Z., 2016. Deep speech 2 : end-to-end speech recognition in English and Mandarin. In: Proceedings of the 33nd International Conference on Machine Learning, ICML.

Anderson, P., He, X., Buehler, C., Teney, D., Johnson, M., Gould, S., Zhang, L., 2018. Bottom-up and top-down attention for image captioning and visual question answering. In: IEEE Conference on Computer Vision and Pattern Recognition, CVPR.

Andreas, J., Rohrbach, M., Darrell, T., Klein, D., 2016a. Learning to Compose Neural Networks for Question Answering, in: NAACL HLT 2016, the 2016 Conference of the North American Chapter of the Association for Computational Linguistics: Human Language Technologies, the Association for Computational Linguistics.

Andreas, J., Rohrbach, M., Darrell, T., Klein, D., 2016b. Neural module networks. In: IEEE Conference on Computer Vision and Pattern Recognition, CVPR, IEEE Computer Society.

Antoniou, G., Williams, M., 1997. Nonmonotonic Reasoning. MIT Press.

Bahdanau, D., Cho, K., Bengio, Y., 2015. Neural machine translation by jointly learning to align and translate. In: 3rd International Conference on Learning Representations, ICLR.

Barclay, J.R., Bransford, J.D., Franks, J.J., McCarrell, N.S., Nitsch, K., 1974. Comprehension and semantic flexibility. J. Verb. Learn. Verb. Behav. 13, 471–481.

Barsalou, L.W., 2008. Grounded cognition. Annu. Rev. Psychol. 59, 617–645.

Battaglia, P.W., Hamrick, J.B., Bapst, V., Sanchez-Gonzalez, A., Zambaldi, V.F., Malinowski, M., Tacchetti, A., Raposo, D., Santoro, A., Faulkner, R., Gülçehre, Ç., Song, H.F., Ballard, A.J., Gilmer, J., Dahl, G.E., Vaswani, A., Allen, K.R., Nash, C., Langston, V., Dyer, C., Heess, N., Wierstra, D., Kohli, P., Botvinick, M., Vinyals, O., Li, Y., Pascanu, R., 2018. Relational Inductive Biases, Deep Learning, and Graph Networks. CoRR abs/1806.01261.

Bengio, Y., 2017. The Consciousness Prior. CoRR abs/1709.08568.

Bengio, S., Heigold, G., 2014. Word embeddings for speech recognition. In: Proceedings of INTERSPEECH 2014, 15th Annual Conference of the International Speech Communication Association.

Bengio, Y., Ducharme, R., Vincent, P., Jauvin, C., 2003. A neural probabilistic language model. J. Mach. Learn. Res. 3, 1137–1155.

Bengio, Y., Marcus, G., Boucher, V., 2019. Yoshua Bengio and Gary Marcus on the best way forward for AI. https://medium.com/@Montreal.AI/ transcript-of-the-ai-debate-1e098eeb8465.

Besold, T.R., d'Avila Garcez, A.S., Bader, S., Bowman, H., Domingos, P.M., Hitzler, P., Kühnberger, K., Lamb, L.C., Lowd, D., Lima, P.M.V., de Penning, L., Pinkas, G., Poon, H., Zaverucha, G., 2017. Neural-symbolic Learning and Reasoning: A Survey and Interpretation. CoRR abs/1711.03902.

Bisk, Y., Holtzman, A., Thomason, J., Andreas, J., Bengio, Y., Chai, J., Lapata, M., Lazaridou, A., May, J., Nisnevich, A., Pinto, N., Turian, J.P., 2020. Experience Grounds Language. CoRR abs/2004.10151.

Bransford, J.D., Johnson, M.K., 1972. Contextual prerequisites for understanding: some investigations of comprehension and recall. J. Verb. Learn. Verb. Behav. 11, 717–726.

Brendel, W., Bethge, M., 2018. Approximating CNNs with bag-of-local-features models works surprisingly well on imagenet. In: 7th International Conference on Learning Representations, ICLR.

Brown, T.B., Mann, B., Ryder, N., Subbiah, M., Kaplan, J., Dhariwal, P., Neelakantan, A., Shyam, P., Sastry, G., Askell, A., Agarwal, S., Herbert-Voss, A., Krueger, G., Henighan, T., Child, R., Ramesh, A., Ziegler, D.M., Wu, J., Winter, C., Hesse, C., Chen, M., Sigler, E., Litwin, M., Gray, S., Chess, B., Clark, J., Berner, C., McCandlish, S., Radford, A., Sutskever, I., Amodei, D., 2020. Language models are few-shot learners. In: Advances in Neural Information Processing Systems 33: Annual Conference on Neural Information Processing Systems.

Burgess, C.P., Matthey, L., Watters, N., Kabra, R., Higgins, I., Botvinick, M., Lerchner, A., 2019. MONet: Unsupervised Scene Decomposition and Representation. CoRR abs/1901.11390.

Calandra, R., Owens, A., Jayaraman, D., Lin, J., Yuan, W., Malik, J., Adelson, E.H., Levine, S., 2018. More than a feeling: learning to grasp and regrasp using vision and touch. IEEE Robotics and Automation Letters 3, 3300–3307.

Carion, N., Massa, F., Synnaeve, G., Usunier, N., Kirillov, A., Zagoruyko, S., 2020. End-to-End Object Detection with Transformers. CoRR abs/2005.12872.

Chan, W., Jaitly, N., Le, Q., Vinyals, O., 2016. Listen, attend and spell: a neural network for large vocabulary conversational speech recognition. In: 2016 IEEE International Conference on Acoustics, Speech and Signal Processing (ICASSP), IEEE.

Chandak, Y., Theocharous, G., Kostas, J., Jordan, S.M., Thomas, P.S., 2019. Learning action representations for reinforcement learning. In: Proceedings of the 36th International Conference on Machine Learning, ICML.

Chen, Y.C., Li, L., Yu, L., Kholy, A.E., Ahmed, F., Gan, Z., Cheng, Y., Liu, J., 2019. UNITER: Learning Universal Image-Text Representations. CoRR abs/1909.11740.

Chen, M., Radford, A., Child, R., Wu, J., Jun, H., Dhariwal, P., Luan, D., Sutskever, I., 2020. Generative pretraining from pixels. In: Proceedings of the 37th International Conference on Machine Learning.

Child, R., Gray, S., Radford, A., Sutskever, I., 2019. Generating Long Sequences with Sparse Transformers. CoRR abs/1904.10509.

Chollet, F., 2019. On the Measure of Intelligence. CoRR abs/1911.01547.

Chomsky, N., 1959. A review of bf skinner's verbal behavior. Language 35, 26–58.

Chomsky, N., 1965. Aspects of the Theory of Syntax. MIT Press.

Chomsky, N., Keyser, S.J., et al., 1988. Language and Problems of Knowledge: the Managua Lectures. MIT press.

Chung, Y., Hsu, W., Tang, H., Glass, J.R., 2019. An unsupervised autoregressive model for speech representation learning. In: Kubin, G., Kacic, Z. (Eds.), Interspeech 2019, 20th Annual Conference of the International Speech Communication Association.

Ciresan, D.C., Meier, U., Masci, J., Gambardella, L.M., Schmidhuber, J., 2011. Flexible, high performance convolutional neural networks for image classification. In: Proceedings of the 22nd International Joint Conference on Artificial Intelligence IJCAI, IJCAI/AAAI.

Collell, G., Zhang, T., Moens, M.F., 2017. Imagined visual representations as multimodal embeddings. In: Proceedings of the Thirty-First AAAI Conference on Artificial Intelligence.

Collell, G., Van Gool, L., Moens, M.F., 2018. Acquiring common sense spatial knowledge through implicit spatial templates. In: Proceedings of the Thirty-Second AAAI Conference on Artificial Intelligence.

Collins, A.M., Loftus, E.F., 1975. A spreading-activation theory of semantic processing. Psychol. Rev. 82, 407–428.

Conneau, A., Kruszewski, G., Lample, G., Barrault, L., Baroni, M., 2018. What you can cram into a single vector: probing sentence embeddings for linguistic properties, in: proceedings of the 56th Annual Meeting of the Association for Computational Linguistics. ACLPPinforma 1. Long Papers.

Cooper, G.F., 1990. The computational complexity of probabilistic inference using bayesian belief networks. Artif. Intell. 42, 393–405.

Cramer, J., Wu, H., Salamon, J., Bello, J.P., 2019. Look, listen, and learn more: design choices for deep audio embeddings. In: IEEE International Conference on Acoustics, Speech and Signal Processing, ICASSP.

Cranmer, M., Sanchez-Gonzalez, A., Battaglia, P., Xu, R., Cranmer, K., Spergel, D., Ho, S., 2020. Discovering Symbolic Models from Deep Learning with Inductive Biases. CoRR abs/2006.11287.

Crevier, D., 1993. AI: the Tumultuous History of the Search for Artificial Intelligence. Basic Books, Inc.

Cuturi, M., Teboul, O., Vert, J.P., 2019. Differentiable ranking and sorting using optimal transport. In: Advances in Neural Information Processing Systems 32: Annual Conference on Neural Information Processing Systems.

Deschacht, K., De Belder, J., Moens, M.F., 2012. The latent words language model. Comput. Speech Lang 26, 384–409.

Devlin, J., Chang, M., Lee, K., Toutanova, K., 2019. BERT: pre-training of deep bidirectional transformers for language understanding. In: Proceedings of the 2019 Conference of the North American Chapter of the Association for Computational Linguistics: Human Language Technologies, NAACL-HLT, vol. 1. Association for Computational Linguistics (Long and Short Papers).

Dulac-Arnold, G., Evans, R., Sunehag, P., Coppin, B., 2015. Deep Reinforcement Learning in Large Discrete Action Spaces. CoRR abs/1512.07679.

d'Avila Garcez, A.S., Lamb, L.C., Gabbay, D.M., 2009. Neural-Symbolic Cognitive Reasoning. Cognitive Technologies, Springer.

Elman, J.L., 1990. Finding structure in time. Cognit. Sci. 14, 179–211.

Engelcke, M., Kosiorek, A.R., Jones, O.P., Posner, I., 2020. GENESIS: generative scene inference and sampling with object-centric latent representations. In: 8th International Conference on Learning Representations, ICLR.

Eslami, S.M.A., Heess, N., Weber, T., Tassa, Y., Szepesvari, D., Kavukcuoglu, K., Hinton, G.E., 2016. Attend, infer, repeat: fast scene understanding with generative models. In: Advances in Neural Information Processing Systems 29: Annual Conference on Neural Information Processing Systems.

Ferrand, L., New, B., 2003. Semantic and associative priming in the mental lexicon. Ment. Lexicon: Some words to talk about words 25–43.

Firth, J.R., 1957. A synopsis of linguistic theory 1930-55 1952–59, 1–32.

Fodor, J.A., 1975. The Language of Thought. Harvard University Press.

Fodor, J.A., Pylyshyn, Z.W., et al., 1988. Connectionism and cognitive architecture: a critical analysis. Cognition 28, 3–71.

Foss, D.J., 1982. A discourse on semantic priming. Cognit. Psychol. 14, 590–607.

Frome, A., Corrado, G.S., Shlens, J., Bengio, S., Dean, J., Ranzato, M., Mikolov, T., 2013. Devise: a deep visual-semantic embedding model. In: Advances in Neural Information Processing Systems 26: 27th Annual Conference on Neural Information Processing Systems.

Gao, Y., Hendricks, L.A., Kuchenbecker, K.J., Darrell, T., 2016. Deep learning for tactile understanding from visual and haptic data. In: 2016 IEEE International Conference on Robotics and Automation (ICRA), IEEE.

Geirhos, R., Jacobsen, J., Michaelis, C., Zemel, R.S., Brendel, W., Bethge, M., Wichmann, F.A., 2020. Shortcut Learning in Deep Neural Networks. CoRR abs/2004.07780.

Glenberg, A.M., Kaschak, M.P., 2002. Grounding language in action. Psychonomic Bull. Rev. 9, 558–565.

Goldberg, Y., 2019. Assessing BERT's Syntactic Abilities. CoRR abs/1901.05287.

Goodfellow, I., Bengio, Y., Courville, A.C., 2016. Deep Learning. Adaptive Computation and Machine Learning. MIT Press.

Graves, A., Fernández, S., Gomez, F.J., Schmidhuber, J., 2006. Connectionist temporal classification: labelling unsegmented sequence data with recurrent neural networks. In: Cohen, W.W., Moore, A.W. (Eds.), Machine Learning, Proceedings of the Twenty-Third International Conference (ICML).

Graves, A., Mohamed, A., Hinton, G.E., 2013. Speech recognition with deep recurrent neural networks. In: IEEE International Conference on Acoustics, Speech and Signal Processing, ICASSP.

Graves, A., Wayne, G., Danihelka, I., 2014. Neural Turing Machines. CoRR abs/1410.5401.

Graves, A., Wayne, G., Reynolds, M., Harley, T., Danihelka, I., Grabska-Barwinska, A., Colmenarejo, S.G., Grefenstette, E., Ramalho, T., Agapiou, J.P., Badia, A.P., Hermann, K.M., Zwols, Y., Ostrovski, G., Cain, A., King, H., Summerfield, C., Blunsom, P., Kavukcuoglu, K., Hassabis, D., 2016. Hybrid computing using a neural network with dynamic external memory. Nature 538, 471–476.

Gulcehre, C., Denil, M., Malinowski, M., Razavi, A., Pascanu, R., Hermann, K.M., Battaglia, P., Bapst, V., Raposo, D., Santoro, A., de Freitas, N., 2019. Hyperbolic attention networks. In: 7th International Conference on Learning Representations, ICLR.

Gülçehre, Ç., Chandar, S., Cho, K., Bengio, Y., 2018. Dynamic neural turing machine with continuous and discrete addressing schemes. Neural Comput. 30.

Hannun, A.Y., Case, C., Casper, J., Catanzaro, B., Diamos, G., Elsen, E., Prenger, R., Satheesh, S., Sengupta, S., Coates, A., Ng, A.Y., 2014. Deep Speech: Scaling up End-To-End Speech Recognition. CoRR abs/1412.5567.

Hao, K., 2020. We Analyzed 16,625 Papers to Figure Out where Ai Is Headed Next. URL: https://www.technologyreview.com/2019/01/25/1436/we-analyzed-16625-papers-to-figure-out-where-ai-is-headed-next/.

Harris, Z.S., 1954. Distributional structure. Word 10, 146–162.

Hasson, U., Nastase, S.A., Goldstein, A., 2020. Direct fit to nature: an evolutionary perspective on biological and artificial neural networks. Neuron 105, 416–434.

He, K., Zhang, X., Ren, S., Sun, J., 2016. Deep residual learning for image recognition. In: IEEE Conference on Computer Vision and Pattern Recognition, CVPR.

Heess, N., Tarlow, D., Winn, J.M., 2013. Learning to pass expectation propagation messages. Advances in Neural Information Processing Systems 26: 27th Annual Conference on Neural Information Processing Systems 3219–3227.

Hewitt, J., Manning, C.D., 2019. A structural probe for finding syntax in word representations. In: Proceedings of the 2019 Conference of the North American Chapter of the Association for Computational Linguistics: Human Language Technologies, NAACL-HLT 2019, vol. 1. Association for Computational Linguistics (Long and Short Papers).

Higgins, I., Matthey, L., Pal, A., Burgess, C., Glorot, X., Botvinick, M., Mohamed, S., Lerchner, A., 2017. Beta-vae: learning basic visual concepts with a constrained variational framework. In: 5th International Conference on Learning Representations, ICLR.

Hill, F., Korhonen, A., 2014. Learning abstract concept embeddings from multi-modal data: since you probably can't see what i mean. In: Proceedings of the 2014 Conference on Empirical Methods in Natural Language Processing (EMNLP).

Hinton, G.E., Sejnowski, T.J., 1983. Optimal perceptual inference. In: Proceedings of the IEEE Conference on Computer Vision and Pattern Recognition.

Hobbes, T., 1914. Leviathan. JM Dent.

Hopfield, J.J., 1982. Neural networks and physical systems with emergent collective computational abilities. Proc. Natl. Acad. Sci. Unit. States Am. 79, 2554–2558.

Howard, J., Ruder, S., 2018. Universal language model fine-tuning for text classification, in: proceedings of the 56th annual meeting of the association for computational linguistics. ACLPPinforma 1. Long Papers.

Hu, R., Andreas, J., Rohrbach, M., Darrell, T., Saenko, K., 2017. Learning to reason: end-to-end module networks for visual question answering. In: IEEE International Conference on Computer Vision, ICCV, IEEE Computer Society.

Hu, R., Andreas, J., Darrell, T., Saenko, K., 2018. Explainable neural computation via stack neural module networks. In: Proceedings of the European Conference on Computer Vision (ECCV).

Hudson, D.A., Manning, C.D., 2018. Compositional attention networks for machine reasoning. In: 6th International Conference on Learning Representations, ICLR.

Hudson, D.A., Manning, C.D., 2019a. GQA: a new dataset for real-world visual reasoning and compositional question answering. In: IEEE Conference on Computer Vision and Pattern Recognition, CVPR.

Hudson, D.A., Manning, C.D., 2019b. Learning by abstraction: the neural state machine. In: Advances in Neural Information Processing Systems 32: Annual Conference on Neural Information Processing Systems.

Hupkes, D., Dankers, V., Mul, M., Bruni, E., 2020. Compositionality decomposed: how do neural networks generalise? J. Artif. Intell. Res. 67, 757–795.

Huth, A.G., Nishimoto, S., Vu, A.T., Gallant, J.L., 2012. A continuous semantic space describes the representation of thousands of object and action categories across the human brain. Neuron 76, 1210–1224.

Huth, A.G., De Heer, W.A., Griffiths, T.L., Theunissen, F.E., Gallant, J.L., 2016. Natural speech reveals the semantic maps that tile human cerebral cortex. Nature 532, 453–458.

Jang, E., Gu, S., Poole, B., 2017. Categorical reparameterization with gumbel-softmax. In: 5th International Conference on Learning Representations, ICLR.

Jansen, A., Plakal, M., Pandya, R., Ellis, D.P., Hershey, S., Liu, J., Moore, R.C., Saurous, R.A., 2018. Unsupervised learning of semantic audio representations. In: 2018 IEEE International Conference on Acoustics, Speech and Signal Processing (ICASSP), IEEE.

Jean, N., Wang, S., Samar, A., Azzari, G., Lobell, D., Ermon, S., 2019. Tile2vec: unsupervised representation learning for spatially distributed data. In: The Thirty-Third AAAI Conference on Artificial Intelligence, AAAI.

Jo, J., Bengio, Y., 2017. Measuring the Tendency of CNNs to Learn Surface Statistical Regularities. CoRR abs/1711.11561.

Johnson, J., Hariharan, B., van der Maaten, L., Fei-Fei, L., Lawrence Zitnick, C., Girshick, R., 2017a. Clevr: a diagnostic dataset for compositional language and elementary visual reasoning. In: IEEE Conference on Computer Vision and Pattern Recognition, CVPR.

Johnson, J., Hariharan, B., van der Maaten, L., Hoffman, J., Fei-Fei, L., Zitnick, C.L., Girshick, R.B., 2017b. Inferring and executing programs for visual reasoning. In: IEEE International Conference on Computer Vision, ICCV, IEEE Computer Society.

Johnson, J., Gupta, A., Fei-Fei, L., 2018. Image generation from scene graphs. In: IEEE Conference on Computer Vision and Pattern Recognition, CVPR.

Kaggle. Abstraction and reasoning challenge leaderboard. https://www.kaggle.com/c/abstraction-and-reasoning-challenge/leaderboard.

Kharitonov, E., Chaabouni, R., 2020. What They Do when in Doubt: a Study of Inductive Biases in Seq2seq Learners. CoRR abs/2006.14953.

Kiela, D., Bottou, L., 2014. Learning image embeddings using convolutional neural networks for improved multi-modal semantics. In: Proceedings of the 2014 Conference on Empirical Methods in Natural Language Processing (EMNLP).

Kiela, D., Bulat, L., Clark, S., 2015. Grounding semantics in olfactory perception. In: Proceedings of the 53rd Annual Meeting of the Association for Computational Linguistics and the 7th International Joint Conference on Natural Language Processing (Volume 2: Short Papers).

Kim, H., Mnih, A., 2018. Disentangling by factorising. In: Proceedings of the 35th International Conference on Machine Learning, ICML.

Kingma, D.P., Welling, M., 2014. Auto-encoding Variational Bayes, in: 2nd International Conference on Learning Representations, ICLR.

Kipf, T.N., Fetaya, E., Wang, K., Welling, M., Zemel, R.S., 2018. Neural relational inference for interacting systems. In: Proceedings of the 35th International Conference on Machine Learning, ICML.

Kipf, T.N., van der Pol, E., Welling, M., 2020. Contrastive Learning of Structured World Models, in: 8th International Conference on Learning Representations, ICLR.

Klein, J., 1989. A Commentary on Plato's Meno. University of Chicago Press.

Krizhevsky, A., Sutskever, I., Hinton, G.E., 2012. Imagenet classification with deep convolutional neural networks. In: Advances in Neural Information Processing Systems 25: Annual Conference on Neural Information Processing Systems.

Lake, B.M., Baroni, M., 2018. Generalization without systematicity: on the compositional skills of sequence-to-sequence recurrent networks. In: Proceedings of the 35th International Conference on Machine Learning, ICML.

Lake, B.M., Salakhutdinov, R., Tenenbaum, J.B., 2015. Human-level concept learning through probabilistic program induction. Science 350, 1332–1338.

Lake, B.M., Ullman, T.D., Tenenbaum, J.B., Gershman, S.J., 2017. Building machines that learn and think like people. Behav. Brain Sci. 40.

Lample, G., Charton, F., 2020. Deep learning for symbolic mathematics. In: 8th International Conference on Learning Representations, ICLR.

Landauer, T., Dumais, S., 1997. A solution to plato's problem: the latent semantic analysis theory of acquisition, induction, and representation of knowledge. Psychol. Rev. 104, 211–240.

Lapuschkin, S., Wäldchen, S., Binder, A., Montavon, G., Samek, W., Müller, K.R., 2019. Unmasking clever hans predictors and assessing what machines really learn. Nat. Commun. 10, 1–8.

Lazaridou, A., Baroni, M., et al., 2015. Combining language and vision with a multimodal skip-gram model. In: Proceedings of the 2015 Conference of the North American Chapter of the Association for Computational Linguistics: Human Language Technologies.

LeCun, Y., Boser, B.E., Denker, J.S., Henderson, D., Howard, R.E., Hubbard, W.E., Jackel, L.D., 1989. Backpropagation applied to handwritten zip code recognition. Neural Comput. 1, 541–551.

Leeuwenberg, A., Moens, M.F., 2018. Temporal information extraction by predicting relative time-lines. In: Proceedings of the 2018 Conference on Empirical Methods in Natural Language Processing.

Leeuwenberg, A., Moens, M.F., 2020. Towards extracting absolute event timelines from English clinical reports. IEEE/ACM Transactions on Audio, Speech, and Language Processing 28, 2710–2719.

Li, G., Duan, N., Fang, Y., Gong, M., Jiang, D., Zhou, M., 2020. Unicoder-vl: a universal encoder for vision and language by cross-modal pre-training. In: The Thirty-Fourth AAAI Conference on Artificial Intelligence, AAAI.

Little, W.A., 1974. The existence of persistent states in the brain. Math. Biosci. 19, 101–120.

Liu, L., Ouyang, W., Wang, X., Fieguth, P., Chen, J., Liu, X., Pietikäinen, M., 2020. Deep learning for generic object detection: a survey. Int. J. Comput. Vis. 128, 261–318.

Locatello, F., Bauer, S., Lucic, M., Rätsch, G., Gelly, S., Schölkopf, B., Bachem, O., 2019. Challenging common assumptions in the unsupervised learning of disentangled representations. In: Proceedings of the 36th International Conference on Machine Learning, ICML.

Locatello, F., Weissenborn, D., Unterthiner, T., Mahendran, A., Heigold, G., Uszkoreit, J., Dosovitskiy, A., Kipf, T., 2020. Object-centric learning with slot attention. In: Advances in Neural Information Processing Systems 33: Annual Conference on Neural Information Processing Systems.

Loula, J., Baroni, M., Lake, B., 2018. Rearranging the familiar: testing compositional generalization in recurrent networks. In: Proceedings of the Workshop: Analyzing and Interpreting Neural Networks for NLP. BlackboxNLP@EMNLP.

Lu, J., Batra, D., Parikh, D., Lee, S., 2019. VilBERT: pretraining task-agnostic visiolinguistic representations for vision-and-language tasks. In: Advances in Neural Information Processing Systems 32: Annual Conference on Neural Information Processing Systems.

Lu, K., Grover, A., Abbeel, P., Mordatch, I., 2021. Pretrained Transformers as Universal Computation Engines. CoRR abs/2103.05247.

Maddison, C.J., Mnih, A., Teh, Y.W., 2017. The concrete distribution: a continuous relaxation of discrete random variables. 5th International Conference on Learning Representations, ICLR.

Manhaeve, R., Dumancic, S., Kimmig, A., Demeester, T., Raedt, L.D., 2018. Deepproblog: neural probabilistic logic programming. In: Advances in Neural Information Processing Systems 31: Annual Conference on Neural Information Processing Systems.

Mao, J., Gan, C., Kohli, P., Tenenbaum, J.B., Wu, J., 2019. The neuro-symbolic concept learner: interpreting scenes, words, and sentences from natural supervision. In: 7th International Conference on Learning Representations, ICLR.

Marcinkevics, R., Vogt, J.E., 2020. Interpretability and Explainability: A Machine Learning Zoo Mini-Tour. CoRR abs/2012.01805.

Marcus, G., 2001. The Algebraic Mind: Integrating Connectionism and Cognitive Science. MIT press.

Marcus, G., 2018. Deep Learning: A Critical Appraisal. CoRR abs/1801.00631.

Marcus, G., 2020. The Next Decade in AI: Four Steps towards Robust Artificial Intelligence. CoRR abs/2002.06177.

Martin-Maroto, F., de Polavieja, G.G., 2018. Algebraic Machine Learning. CoRR abs/1803.05252.

Mascharka, D., Tran, P., Soklaski, R., Majumdar, A., 2018. Transparency by design: closing the gap between performance and interpretability in visual reasoning,. In: IEEE Conference on Computer Vision and Pattern Recognition, CVPR.

McClelland, J.L., Hill, F., Rudolph, M., Baldridge, J., Schütze, H., 2019. Extending Machine Language Models toward Human-Level Understanding. CoRR abs/1912.05877.

McClelland, J.L., Hill, F., Rudolph, M., Baldridge, J., Schütze, H., 2020. Placing language in an integrated understanding system: Next steps toward human-level comprehension in neural language models. Proc. Natl. Acad. Sci. U.S.A. 117, 25966–25974.

McCoy, R.T., Frank, R., Linzen, T., 2020. Does syntax need to grow on trees? sources of hierarchical inductive bias in sequence-to-sequence networks. Transactions of the Association for Computational Linguistics 8, 125–140.

McRae, K., De Sa, V.R., Seidenberg, M.S., 1997. On the nature and scope of featural representations of word meaning. J. Exp. Psychol. Gen. 126, 99.

Meyer, D., Schvaneveldt, R., 1971. Facilitation in recognizing pairs of words: evidence of a dependence between retrieval operations. J. Exp. Psychol. 90, 227–234.

Mikolov, T., Chen, K., Corrado, G., Dean, J., 2013a. Efficient estimation of word representations in vector space. In: 1st International Conference on Learning Representations, ICLR, Workshop Track Proceedings.

Mikolov, T., Sutskever, I., Chen, K., Corrado, G.S., Dean, J., 2013b. Distributed representations of words and phrases and their compositionality. Advances in Neural Information Processing Systems 26: 27th Annual Conference on Neural Information Processing Systems.

Minsky, M., 1991. Logical versus analogical or symbolic versus connectionist or neat versus scruffy. AI Mag. 12, 34–51.

Minsky, M., Papert, S., 1969. Perceptrons - an Introduction to Computational Geometry. MIT Press.

Mitchell, T.M., Shinkareva, S.V., Carlson, A., Chang, K.M., Malave, V.L., Mason, R.A., Just, M.A., 2008. Predicting human brain activity associated with the meanings of nouns. Science 320, 1191–1195.

Mitra, B., Diaz, F., Craswell, N., 2017. Learning to match using local and distributed representations of text for web search. In: Proceedings of the 26th International Conference on World Wide Web.

Monett, D., Lewis, C.W.P., Thórisson, K.R., Bach, J., Baldassarre, G., Granato, G., Berkeley, I.S.N., Chollet, F., Crosby, M., Shevlin, H., Sowa, J.F., Laird, J.E., Legg, S., Lindes, P., Mikolov, T., Rapaport, W.J., Rojas, R., Rosa, M., Stone, P., Sutton, R.S., Yampolskiy, R.V., Wang, P., Schank, R.C., Sloman, A., Winfield, A.F.T., 2020. Special issue: "On defining artificial intelligence" - commentaries and author's response. J. Artif. Gen. Intell. 11, 1–100.

Nair, A., Pong, V., Dalal, M., Bahl, S., Lin, S., Levine, S., 2018. Visual reinforcement learning with imagined goals. In: Advances in Neural Information Processing Systems 31: Annual Conference on Neural Information Processing Systems.

Niven, T., Kao, H.Y., 2019. Probing neural network comprehension of natural language arguments. In: Proceedings of the 57th Annual Meeting of the Association for Computational Linguistics.

Pandey, A., Fanuel, M., Schreurs, J., Suykens, J.A., 2020. Disentangled Representation Learning and Generation with Manifold Optimization. CoRR abs/2006.07046.

Parisi, G.I., Kemker, R., Part, J.L., Kanan, C., Wermter, S., 2019. Continual lifelong learning with neural networks: a review. Neural Network. 113, 54–71.

Pavllo, D., Lucchi, A., Hofmann, T., 2020a. Controlling style and semantics in weakly-supervised image generation. Proceedings of the European Conference on Computer Vision (ECCV).

Pavllo, D., Spinks, G., Hofmann, T., Moens, M.F., Lucchi, A., 2020b. Convolutional generation of textured 3D meshes. In: Advances in Neural Information Processing Systems 33: Annual Conference on Neural Information Processing Systems.

Pearl, J., 1989. Probabilistic Reasoning in Intelligent Systems - Networks of Plausible Inference. Morgan Kaufmann Series in Representation and Reasoning. Morgan Kaufmann.

Pennington, J., Socher, R., Manning, C.D., 2014. Glove: global vectors for word representation. In: Proceedings of the 2014 Conference on Empirical Methods in Natural Language Processing (EMNLP).

Peters, M.E., Neumann, M., Iyyer, M., Gardner, M., Clark, C., Lee, K., Zettlemoyer, L., 2018. Deep contextualized word representations. In: Proceedings of the 2018 Conference of the North American Chapter of the Association for Computational Linguistics: Human Language Technologies, NAACL-HLT, vol. 1. Association for Computational Linguistics (Long Papers).

Pinker, S., 1994. The Language Instinct: How the Mind Creates Language. William Morrow and Company.

Radford, A., Narasimhan, K., Salimans, T., Sutskever, I., 2018. Improving language understanding by generative pre-training. Technical Report. OpenAI.

Polyn, S.M., Norman, K.A., Kahana, M.J., 2009. A context maintenance and retrieval model of organizational processes in free recall. Psychol. Rev. 116, 129.

Radford, A., Wu, J., Child, R., Luan, D., Amodei, D., Sutskever, I., 2019. language models are unsupervised multitask learners. Technical Report. OpenAI.

Radford, A., Kim, J.W., Hallacy, C., Ramesh, A., Goh, G., Agarwal, S., Sastry, G., Askell, A., Mishkin, P., Clark, J., Krueger, G., Sutskever, I., 2021. Learning Transferable Visual Models from Natural Language Supervision. CoRR abs/2103.00020.

Rae, J.W., Hunt, J.J., Danihelka, I., Harley, T., Senior, A.W., Wayne, G., Graves, A., Lillicrap, T., 2016. Scaling memory-augmented neural networks with sparse reads and writes. In: Advances in Neural Information Processing Systems 29: Annual Conference on Neural Information Processing Systems.

Raffel, C., Shazeer, N., Roberts, A., Lee, K., Narang, S., Matena, M., Zhou, Y., Li, W., Liu, P.J., 2020. Exploring the limits of transfer learning with a unified text-to-text transformer. J. Mach. Learn. Res. 21, 1–67.

Razavian, A.S., Azizpour, H., Sullivan, J., Carlsson, S., 2014. CNN features off-the-shelf: an astounding baseline for recognition. In: IEEE Conference on Computer Vision and Pattern Recognition, CVPR Workshops.

Rezende, D.J., Mohamed, S., Wierstra, D., 2014. Stochastic backpropagation and approximate inference in deep generative models. In: Proceedings of the 31th International Conference on Machine Learning, ICML.

Rosset, C., 2019. Turing-nlg: A 17-Billion-Parameter Language Model by Microsoft. Microsoft Blog.

Russell, S.J., Norvig, P., 2010. Artificial Intelligence - A Modern Approach, Third International Edition. Pearson Education.

Santoro, A., Raposo, D., Barrett, D.G., Malinowski, M., Pascanu, R., Battaglia, P., Lillicrap, T., 2017. A simple neural network module for relational reasoning. In: Advances in Neural Information Processing Systems 30: Annual Conference on Neural Information Processing Systems.

Satorras, V.G., Welling, M., 2020. Neural Enhanced Belief Propagation on Factor Graphs. CoRR abs/2003.01998.

Satorras, V.G., Welling, M., Akata, Z., 2019. Combining generative and discriminative models for hybrid inference. In: Advances in Neural Information Processing Systems 32: Annual Conference on Neural Information Processing Systems.

Scarselli, F., Gori, M., Tsoi, A.C., Hagenbuchner, M., Monfardini, G., 2008. The graph neural network model. IEEE Trans. Neural Network. 20, 61–80.

Schneider, S., Baevski, A., Collobert, R., Auli, M., 2019. wav2vec: unsupervised pre-training for speech recognition. In: Kubín, G., Kacic, Z. (Eds.), Interspeech 2019, 20th Annual Conference of the International Speech Communication Association, ISCA.

Schütt, K., Sauceda, H., Kindermans, P., Tkatchenko, A., Müller, K., 2018. Schnet-a deep learning architecture for molecules and materials. J. Chem. Phys. 148, 241722–241722.

Shannon, C.E., 1948. A mathematical theory of communication. The Bell system technical journal 27, 379–423.

Shannon, C.E., Weaver, W., 1949. The Mathematical Theory of Communication. University of Illinois Press.

Silberer, C., Lapata, M., 2014. Learning grounded meaning representations with autoencoders. In: Proceedings of the 52nd Annual Meeting of the Association for Computational Linguistics (Volume 1: Long Papers).

Silver, D., Huang, A., Maddison, C.J., Guez, A., Sifre, L., van den Driessche, G., Schrittwieser, J., Antonoglou, I., Panneershelvam, V., Lanctot, M., Dieleman, S., Grewe, D., Nham, J., Kalchbrenner, N., Sutskever, I., Lillicrap, T.P., Leach, M., Kavukcuoglu, K., Graepel, T., Hassabis, D., 2016. Mastering the game of go with deep neural networks and tree search. Nature 529, 484–489.

Silver, D., Schrittwieser, J., Simonyan, K., Antonoglou, I., Huang, A., Guez, A., Hubert, T., Baker, L., Lai, M., Bolton, A., Chen, Y., Lillicrap, T.P., Hui, F., Sifre, L., van den Driessche, G., Graepel, T., Hassabis, D., 2017. Mastering the game of go without human knowledge. Nature 550, 354–359.

Simonyan, K., Zisserman, A., 2015. Very deep convolutional networks for large-scale image recognition. In: 3rd International Conference on Learning Representations, ICLR.

Smolensky, P., 1986. Information Processing in Dynamical Systems: Foundations of Harmony Theory. MIT Press, Cambridge, MA, USA, pp. 194–281.

Socher, R., Ganjoo, M., Manning, C.D., Ng, A., 2013. Zero-shot learning through cross-modal transfer. In: Advances in Neural Information Processing Systems 26: 27th Annual Conference on Neural Information Processing Systems.

Sperber, D., Wilson, D., 1986. Relevance: Communication and Cognition, vol. 142. Harvard University Press Cambridge, MA.

Spinks, G., Moens, M.F., 2020. Structured (de)composable representations trained with neural networks. Computers 9, 79.

Stanovich, K.E., West, R.F., 1983. On priming by a sentence context. J. Exp. Psychol. Gen. 112, 1.

Su, W., Zhu, X., Cao, Y., Li, B., Lu, L., Wei, F., Dai, J., 2019. VL-BERT: pre-training of generic visual-linguistic representations. In: 8th International Conference on Learning Representations, ICLR.

Sukhbaatar, S., Szlam, A., Weston, J., Fergus, R., 2015. End-to-end memory networks. In: Advances in Neural Information Processing Systems 28: Annual Conference on Neural Information Processing Systems.

Sun, R., Bookman, L.A., 1994. Computational Architectures Integrating Neural and Symbolic Processes: A Perspective on the State of the Art, vol. 292. Springer Science & Business Media.

Sun, C., Myers, A., Vondrick, C., Murphy, K., Schmid, C., 2019. VideoBERT: a joint model for video and language representation learning. In: IEEE/CVF International Conference on Computer Vision, ICCV.

Sung, F., Yang, Y., Zhang, L., Xiang, T., Torr, P.H., Hospedales, T.M., 2018. Learning to compare: relation network for few-shot learning. In: IEEE Conference on Computer Vision and Pattern Recognition, CVPR.

Sutskever, I., Vinyals, O., Le, Q.V., 2014. Sequence to sequence learning with neural networks. In: Advances in Neural Information Processing Systems 27: Annual Conference on Neural Information Processing Systems.

Sutton, R.S., McAllester, D.A., Singh, S.P., Mansour, Y., 1999. Policy gradient methods for reinforcement learning with function approximation. In: Advances in Neural Information Processing Systems 12. The MIT Press.

Tan, H., Bansal, M., 2020. Vokenization: improving language understanding with contextualized, visual-grounded supervision. In: Proceedings of the 2020 Conference on Empirical Methods in Natural Language Processing, EMNLP. Association for Computational Linguistics.

Tang, K., Niu, Y., Huang, J., Shi, J., Zhang, H., 2020. Unbiased scene graph generation from biased training. In: IEEE/CVF Conference on Computer Vision and Pattern Recognition, CVPR.

Tenney, I., Xia, P., Chen, B., Wang, A., Poliak, A., McCoy, R.T., Kim, N., Durme, B.V., Bowman, S.R., Das, D., Pavlick, E., 2019. What do you learn from context? probing for sentence structure in contextualized word representations. In: 7th International Conference on Learning Representations, ICLR.

van den Oord, A., Li, Y., Vinyals, O., 2018. Representation Learning with Contrastive Predictive Coding. CoRR abs/1807.03748.

Vaswani, A., Shazeer, N., Parmar, N., Uszkoreit, J., Jones, L., Gomez, A.N., Kaiser, Ł., Polosukhin, I., 2017. Attention is all you need. In: Advances in Neural Information Processing Systems 30: Annual Conference on Neural Information Processing Systems.

Vedantam, R., Fischer, I., Huang, J., Murphy, K., 2018. Generative models of visually grounded imagination. In: 6th International Conference on Learning Representations, ICLR.

Veličković, P., Cucurull, G., Casanova, A., Romero, A., Liò, P., Bengio, Y., 2018. Graph attention networks. In: 6th International Conference on Learning Representations, ICLR.

Veličković, P., Fedus, W., Hamilton, W.L., Liò, P., Bengio, Y., Hjelm, R.D., 2019. Deep graph infomax. In: 7th International Conference on Learning Representations, ICLR.

Vinyals, O., Fortunato, M., Jaitly, N., 2015a. Pointer networks. In: Advances in Neural Information Processing Systems 28: Annual Conference on Neural Information Processing Systems.

Vinyals, O., Toshev, A., Bengio, S., Erhan, D., 2015b. Show and tell. In: A Neural Image Caption Generator, in: IEEE Conference on Computer Vision and Pattern Recognition, CVPR.

Vlastelica, M.P., Paulus, A., Musil, V., Martius, G., Rolínek, M., 2020. Differentiation of blackbox combinatorial solvers. In: 8th International Conference on Learning Representations, ICLR.

von Humboldt, W., von Humboldt, W.F., Losonsky, M., Heath, P., Yao, X., von, H.W., Ameriks, K., Clarke, D.M., 1999. Humboldt: 'On Language': on the Diversity of Human Language Construction and its Influence on the Mental Development of the Human Species. Cambridge Texts in the History of Philosophy. Cambridge University Press.

Wang, P., 2019. On defining artificial intelligence. J. Artif. Gen. Intell. 10, 1–37.

Watters, N., Matthey, L., Bosnjak, M., Burgess, C.P., Lerchner, A., 2019. COBRA: Data-Efficient Model-Based RL through Unsupervised Object Discovery and Curiosity-Driven Exploration. CoRR abs/1905.09275.

Wei, X., Croft, W.B., 2006. Lda-based document models for ad-hoc retrieval. In: SIGIR 2006: Proceedings of the 29th Annual International ACM SIGIR Conference on Research and Development in Information Retrieval, ACM.

Weston, J., Chopra, S., Bordes, A., 2015. Memory networks. In: 3rd International Conference on Learning Representations, ICLR.

Weston, J., Bordes, A., Chopra, S., Mikolov, T., 2016. Towards ai-complete question answering: a set of prerequisite toy tasks. In: 4th International Conference on Learning Representations, ICLR.

Williams, R.J., 1992. Simple statistical gradient-following algorithms for connectionist reinforcement learning. Mach. Learn. 8, 229–256.

Xie, S., Sun, C., Huang, J., Tu, Z., Murphy, K., 2018. Rethinking spatiotemporal feature learning: speed-accuracy trade-offs in video classification. In: Proceedings of the European Conference on Computer Vision (ECCV), Springer.

Xu, Y., Chun, M.M., 2009. Selecting and perceiving multiple visual objects. Trends Cognit. Sci. 13, 167–174.

Xu, K., Ba, J., Kiros, R., Cho, K., Courville, A., Salakhudinov, R., Zemel, R., Bengio, Y., 2015. Show, attend and tell: neural image caption generation with visual attention. In: Proceedings of the 32nd International Conference on Machine Learning, ICML.

Xu, T., Zhang, P., Huang, Q., Zhang, H., Gan, Z., Huang, X., He, X., 2018. AttnGAN: fine-grained text to image generation with attentional generative adversarial networks. In: IEEE Conference on Computer Vision and Pattern Recognition, CVPR, IEEE Computer Society.

Yi, K., Wu, J., Gan, C., Torralba, A., Kohli, P., Tenenbaum, J., 2018. Neural-symbolic VQA: disentangling reasoning from vision and language understanding. In: Advances in Neural Information Processing Systems 31: Annual Conference on Neural Information Processing Systems.

Yonelinas, A.P., Ranganath, C., Ekstrom, A.D., Wiltgen, B.J., 2019. A contextual binding theory of episodic memory: systems consolidation reconsidered. Nat. Rev. Neurosci. 20, 364–375.

Yoon, K., Liao, R., Xiong, Y., Zhang, L., Fetaya, E., Urtasun, R., Zemel, R.S., Pitkow, X., 2018. Inference in probabilistic graphical models by graph neural networks. In: 6th International Conference on Learning Representations, ICLR.

Yu, Z., Yu, J., Cui, Y., Tao, D., Tian, Q., 2019. Deep modular co-attention networks for visual question answering. In: IEEE Conference on Computer Vision and Pattern Recognition, CVPR.

Yuan, W., Wang, S., Dong, S., Adelson, E., 2017. Connecting look and feel: associating the visual and tactile properties of physical materials. In: IEEE Conference on Computer Vision and Pattern Recognition, CVPR.

Zablocki, E., Piwowarski, B., Soulier, L., Gallinari, P., 2018. Learning multi-modal word representation grounded in visual context. Proceedings of the Thirty-Second AAAI Conference on Artificial Intelligence 5626–5633.

Zamani, H., Dehghani, M., Croft, W.B., Learned-Miller, E., Kamps, J., 2018. From neural re-ranking to neural ranking: learning a sparse representation for inverted indexing. In: Proceedings of the 27th ACM International Conference on Information and Knowledge Management, ACM.

Zaremba, W., Sutskever, I., 2015. Reinforcement Learning Neural Turing Machines. CoRR abs/1505.00521.

Zhang, H., Xu, T., Li, H., Zhang, S., Wang, X., Huang, X., Metaxas, D.N., 2017. StackGAN: text to photo-realistic image synthesis with stacked generative adversarial networks. In: IEEE Conference on Computer Vision and Pattern Recognition, CVPR.

Zhou, J., Cui, G., Zhang, Z., Yang, C., Liu, Z., Wang, L., Li, C., Sun, M., 2018. Graph Neural Networks: A Review of Methods and Applications. CoRR abs/1812.08434.

Zipf, G.K., 1935. The Psychology of Language. Houghton-Mifflin.

Łstrokukasiewicz, J., 1951. Aristotle's Syllogistic, from the Standpoint of Modern Formal Logic. Clarendon Press, Oxford, Oxford.